%% file: main_rebuttal.tex
\documentclass[conference]{IEEEtran}
\IEEEoverridecommandlockouts
\usepackage[colorlinks=true, linkcolor=blue, citecolor=blue, urlcolor=blue]{hyperref}
\usepackage{cite}
\usepackage{amsmath,amssymb,amsfonts,amsthm}
\usepackage{enumitem}
\usepackage{graphicx}
\usepackage{textcomp}
\usepackage{xcolor}
\usepackage{algorithm}
\usepackage{algcompatible}
\usepackage{caption}
\usepackage{multirow}
\usepackage{lipsum}
\usepackage{booktabs}
\usepackage{algpseudocode}
\usepackage{booktabs}
\usepackage{siunitx}
\usepackage{array}
\usepackage{makecell}
\usepackage{subcaption} 
\usepackage{pgfplots}
\pgfplotsset{compat=1.18} 
\usepackage{pgfplotstable} 
\usepackage{tikz}
\usepackage{pgfplots}
\pgfplotsset{compat=1.17}
\usetikzlibrary{pgfplots.groupplots}
\usetikzlibrary{pgfplots.fillbetween}
\usepackage{float}
\usepackage{multicol}

\pgfplotsset{
  every nth point/.style={
    x filter/.code={
      \pgfmathparse{int(mod(\coordindex,#1)==0 ? 1 : 0)}
      \ifnum\pgfmathresult=1\relax
      \else
        \def\pgfmathresult{}
      \fi
    }
  }
}

\definecolor{myblue}{RGB}{192,231,252}
\definecolor{myorange}{RGB}{255,219,187}

\newfloat{game}{htbp}{loa}
\floatname{game}{Game}

\newcommand{\rSMAPE}{\makecell[l]{rSMAPE}}
\newcommand{\rSMAPEtext}{rSMAPE}
\newcommand{\spm}[1]{{\scriptsize $\pm$#1}}
\newcommand{\B}[1]{\textbf{#1}}  
\newcolumntype{P}[1]{>{\centering\arraybackslash}p{#1}}  
\newcolumntype{R}[1]{>{\raggedleft\arraybackslash}p{#1}} 
\newcommand{\scl}{\scriptsize}

\def\BibTeX{{\rm B\kern-.05em{\sc i\kern-.025em b}\kern-.08em
    T\kern-.1667em\lower.7ex\hbox{E}\kern-.125emX}}
\begin{document}

\title{Privacy Risks in Time Series Forecasting: \\User- and Record-Level Membership Inference}

\author{
\IEEEauthorblockN{
Nicolas Johansson\IEEEauthorrefmark{3}\IEEEauthorrefmark{1},
Tobias Olsson\IEEEauthorrefmark{3}\IEEEauthorrefmark{1},
Daniel Nilsson\IEEEauthorrefmark{2},
Johan Östman\IEEEauthorrefmark{2},
Fazeleh Hoseini\IEEEauthorrefmark{2}
}
\IEEEauthorblockA{\IEEEauthorrefmark{1}Chalmers University of Technology \quad \IEEEauthorrefmark{2}AI Sweden}
\thanks{\IEEEauthorrefmark{3}These authors contributed equally.}
\thanks{This work was supported by Vinnova, the Swedish innovation agency, under grants 2023-03000 and 2025-03066.}
}

\maketitle

\begin{abstract}
Membership inference attacks (MIAs) aim to determine whether specific data were used to train a model. While extensively studied on classification models, their impact on time series forecasting remains largely unexplored. We address this gap by introducing two new attacks: (i) an adaptation of multivariate LiRA, a state-of-the-art MIA originally developed for classification models, to the time series forecasting setting, and (ii) a novel end-to-end learning approach called Deep Time Series (DTS) attack. We evaluate our attacks in realistic settings on the TUH-EEG and ELD datasets, targeting two strong forecasting architectures, LSTM and the state-of-the-art N-HiTS, under both record- and user-level threat models. Our results show that forecasting models are vulnerable, with user-level attacks often achieving perfect detection. The proposed methods achieve the strongest performance in several settings, establishing new baselines for privacy risk assessment in time series forecasting. Furthermore, the vulnerability increases with longer prediction horizons and smaller training populations, echoing trends observed in large language models.

\end{abstract}


\input{01}
\input{02}

\input{03}

\input{04}

\input{05}

\input{06}

\input{07}
\input{08}

\bibliographystyle{IEEEtran}
\bibliography{references}  
\input{09}

\end{document}

%% file: 01.tex
\section{Introduction}
The integration of machine learning (ML) into real-world applications has significantly expanded the ways in which industries use data for decision-making. Time series forecasting, in particular, is widely applied in domains such as healthcare, finance, and smart infrastructure. However, when deployed in settings that involve sensitive information, these models raise important concerns regarding data privacy.

Modern ML models have been shown to memorize information from their training data~\cite{shokri2017membership}. 
Recent work demonstrates that memorized information can, in some cases, be extracted by adversaries interacting with the model~\cite{ye2022enhanced, zarifzadeh2024lowcost}.
These findings have driven interest in developing empirical methods to assess how much sensitive information a model may inadvertently reveal, commonly referred to as privacy auditing~\cite{murakonda2020ml}.

Membership inference attacks (MIAs) constitute the most fundamental building block of privacy auditing. 
The goal of a MIA is to infer whether a specific data point was part of a model’s training set. 
Although simply inferring membership may not necessary breach privacy, a MIA can be used as a component in more sophisticated attacks, e.g., data reconstruction~\cite{carlini2021extracting, zhang2025position}.
Moreover, a successful MIA implies a degree of memorization or overfitting~\cite{carlini2019secret} that may expose vulnerabilities towards other types of attacks~\cite{salem2023sok}.
Furthermore, MIAs have proven useful in evaluating machine unlearning methods~\cite{di2025adversarial}, establishing lower bounds on the privacy guarantees of differentially-private algorithms~\cite{nasr2021adversary}, and in identifying dataset contamination~\cite{oren2023proving}. 

Despite the extensive study of MIAs across domains such as image classification~\cite{carlini2022first, liu2022membership, zarifzadeh2024lowcost}, text generation~\cite{shi2023detecting, mattern2023membership, duan2024do}, and graph neural networks~\cite{krüger2024publishingneuralnetworksdrug, lassila2025practical}, their implications for time series forecasting models remain largely unexplored. 
A few studies have investigated membership inference in the context of aggregated time series data, where an adversary attempts to determine whether an individual's data was included in a published group-level summary~\cite{pyrgelis2017knock,voyez2022membership}. 
In contrast, our work focuses on predictive models trained with time series data, where the question is whether a given data point was used to train the model.

In time series datasets, data is typically collected continuously from individual users or devices over a period of time. 
To make such data suitable for different machine learning tasks, it is commonly segmented into fixed-length sequences. 
This process results in multiple overlapping or adjacent segments, i.e., training records, derived from the same user. 
Consequently, each user contributes multiple data points to the training set. This is similar to the case of some text analysis tasks, where a single document is partitioned into smaller units, such as paragraphs or sentences~\cite{mireshghallah2022quantifying, meeus2024did}.
In this work, we distinguish MIAs targeting single-record membership from those targeting user-level (multi-record) membership.

At a deeper level, it remains unclear what the appropriate unit of analysis should be in MIAs. Is it sufficient to assess privacy risk at the level of individual records, or should the focus shift to user-level exposure, given that multiple records may collectively reveal the presence of a single user in the training data?
Similar concerns have been raised in the context of large language models (LLMs) where the quantity of interest may vary from sequence level to entire datasets~\cite{meeus2025nowhere}.

MIAs against time series forecasting were recently studied at the record level~\cite{koren2024membership}, demonstrating that unique characteristics of time series, e.g., seasonality and trend, can be exploited by MIAs.
While this study provides valuable insights, it falls short of providing a comprehensive analysis of how various forecasting models respond to MIAs, and omits critical details on the performance of these models, e.g., the degree of overfitting, and the data processing.
Moreover, against classification models, their ensemble method does not perform on par with recent state-of-the-art MIAs such as the Likelihood Ratio Attack (LiRA)~\cite{carlini2022first} or the Robust Membership Inference Attack (RMIA)~\cite{zarifzadeh2024lowcost}.
These attacks may also be effective for time series forecasting: LiRA’s recent extension to LLMs demonstrates how correlations in sequential loss trajectories can be exploited~\cite{rossi2025membership}.

This paper makes the following contributions:\footnote{The code is available at \url{https://github.com/aidotse/LeakPro/tree/SaTML-time-series-mia}.}
\begin{itemize}
    \item We devise two novel attacks: a statistics-based one that builds on multivariate LiRA, and a novel classifier-based one, coined Deep Time Series (DTS), which achieves state-of-the-art performance in the record-level setting.
    
    \item We provide the first systematic investigation of state-of-the-art MIAs in the context of time series forecasting. In particular, we adapt two state-of-the-art membership inference attacks, LiRA and RMIA, originally developed for classification tasks, to the forecasting setting by incorporating time series-specific signals.
   
    \item  We conduct systematic experiments across realistic forecasting architectures, real-world time series datasets, and two adversarial threat models, targeting both record-level and user-level MIAs.

\end{itemize}

%% file: 02.tex
\section{Related Work}

\subsection{Membership Inference Attacks}
Membership inference attacks were first introduced in early work~\cite{shokri2017membership} and have since been extensively studied~\cite{carlini2022first,zarifzadeh2024lowcost,ye2022enhanced,liu2022membership,lassila2025practical,li2024seqmia}. They are commonly divided into statistics-based and classifier-based approaches. Statistics-based methods~\cite{carlini2022first,zarifzadeh2024lowcost,ye2022enhanced,lassila2025practical} compute metrics from the outputs of the target and auxiliary shadow models, which are trained to replicate the target’s behavior. Building on this idea, state-of-the-art variants refine the attack by calibrating thresholds to the difficulty of each sample using shadow models~\cite{carlini2022first,zarifzadeh2024lowcost,lassila2025practical}. A key limitation, however, is that they often fail to distinguish members from non-members when their outputs are too similar~\cite{liu2022membership}.

In contrast, classifier-based attacks~\cite{shokri2017membership} train a separate classifier to distinguish between the target model's behavior on training and non-training inputs. To improve discrimination, some approaches train this classifier on features obtained by mimicking the target model's training process at different stages~\cite{liu2022membership,li2024seqmia}, allowing the attack to differentiate between data points with similar target model outputs.

\subsection{Quantity of Interest in MIAs}
\label{sec:background-quantity}

Most of the MIA literature focuses on individual records. However, in some contexts, such as copyright disputes or dataset contamination, the goal may instead be to infer the membership of a collection of records~\cite{oren2023proving, chen2024copybench}. In the context of text, for instance, quantities of interest may include sequences, paragraphs, documents, or even entire datasets~\cite{meeus2025nowhere}. Earlier works often apply record-level MIAs and then aggregate the results across multiple records to produce a membership score for the collection. This approach has proven more effective than inferring membership for individual records~\cite{mireshghallah2022quantifying, meeus2024did}.

Most existing collection-level studies assume that the adversary has direct access to some of the training records, a setting referred to as user-level membership inference. In contrast, Kandpal et al.~\cite{kandpal2024user} consider a user inference setting, where the attacker does not have access to exact training samples but only to other data points drawn from the same distribution. In our work, we focus on the former setting, user-level membership inference.

\subsection{MIAs on Time Series Forecasting Models}
\label{sec:background-mia-timeseries}
MIAs have primarily been studied in the context of image classification and text generation, while their application to time series models remains underexplored. However, recent work has leveraged time series-specific properties, such as loss trajectories, to design classifier-based attacks~\cite{liu2022membership, li2024seqmia}. For instance, Li et al.~\cite{li2024seqmia} extract features such as fluctuation, correlation, and decline rate from the loss trajectory, which are then used to train a classifier that distinguishes between member and non-member samples.
Similarly, the loss trajectory of text sequences was used together with LiRA to attack LLMs, using more than 400 shadow models~\cite{rossi2025membership}.

Recent work has examined attacks against time series forecasting models~\cite{koren2024membership}.
Their main contribution is the introduction of two time series-specific attack features: trend and seasonality. Seasonality is captured using a 2D Discrete Fourier Transform (2D-DFT), which extracts frequency components across time and variables. The features include Fourier coefficients of the predicted and true sequences, along with their L2 difference. For trend, a low-degree polynomial is fitted to each sequence, and the resulting features are the polynomial coefficients and their L2 difference.

%% file: 03.tex
\section{Preliminaries}
\subsection{Signals as Attack Features}\label{sec:background-signals}
In the context of membership inference against time series models, various signals can be extracted from model outputs to distinguish between training and non-training examples.
We outline the core signals explored in prior work and our study.

\subsubsection{Error-Based Signals}
Standard forecasting metrics such as Mean Squared Error (MSE), Mean Absolute Error (MAE), and Symmetric Mean Absolute Percentage Error (SMAPE) quantify discrepancies between the predicted and true future sequences~\cite{makridakis1993accuracy}. While MSE and MAE are scale-dependent, SMAPE is scale-invariant and bounded between 0 and 1.

\subsubsection{Structural Signals}
Trend and seasonality are commonly used to characterize high-level temporal structure in time series data. Koren et al.~\cite{koren2024membership} leverage these components to capture broader temporal patterns for membership inference attacks, as discussed in Section~\ref{sec:background-mia-timeseries}.

\subsubsection{Representation-Based Signals}
TS2Vec~\cite{yue2022ts2vec} offers a contrastive learning framework that produces rich, multiscale embeddings of time series. Using a shared encoder and hierarchical contrastive loss, TS2Vec learns representations that can generalize across tasks. In our attack context, the L2 distance between the TS2Vec embeddings of predicted and ground-truth sequences acts as a discriminative feature.

\subsection{Time Series Forecasting}
\label{sec:forecasting}

We consider a dataset composed of time series samples collected from $U$ distinct entities, each potentially of different length.  
Let $T_u$ denote the time series associated with entity $u \in [U]$.  
From $T_u$, we construct $n_u$ training records
\begin{equation*}
\mathcal{D}_u = \{ (X_{u,1}, Y_{u,1}), \dots, (X_{u,n_u}, Y_{u,n_u}) \}.
\end{equation*}
Each record $j \in [n_u]$ consists of an input sequence
\begin{equation*}
X_{u,j} = [x_{t_j-L+1}, \dots, x_{t_j}] \in \mathbb{R}^{M \times L},
\end{equation*}
where $L$ is the look-back window length, $M$ is the number of variables, and $t_j$ denotes the index of the last observation in the input sequence, i.e., the time step at the training–label split, and an output sequence (target label)
\begin{equation*}
Y_{u,j} = [y_{t_j+1}, \dots, y_{t_j+H}] \in \mathbb{R}^{M \times H},
\end{equation*}
where $H$ is the prediction horizon.
The complete dataset is given as  
\begin{equation*}
\mathcal{D} = \bigcup_{u \in [U]} \mathcal{D}_u .
\end{equation*}  
The goal is to train a model $f_\theta$, parameterized by $\theta$, to predict $H$ future time steps given a sequence $X \in \mathbb{R}^{M \times L}$, i.e.,
\begin{equation*}
\hat{Y} = f_\theta(X) \in \mathbb{R}^{M \times H}.
\end{equation*}
We refer to the (randomized) training process as $\mathcal{T}$, such that $\theta = \mathcal{T}(\mathcal{D}_{\mathrm{train}})$, where $\mathcal{D}_{\mathrm{train}} \subseteq \mathcal{D}$.

%% file: 04.tex
\section{Threat Model and Setup}

We adopt a game-based formulation of MIAs~\cite{ye2022enhanced, salem2023sok}, modeling the attack as an adversarial game between a challenger and an attacker.
We assume that each entity $u \in [U]$ samples their data from a distribution $\pi_u$. 
Hence, the user dataset $\mathcal{D}_u$ contains records $X_{u,j} \sim \pi_u$ for $j \in [n_u]$.  
We denote a dataset $\mathcal{D}_u$, of size $n_u$, sampled i.i.d.\ from $\pi_u$ as $\mathcal{D}_u \sim \pi_u^{n_u}$.
Moreover, we denote by $\pi$ the distribution over records across all users, referred to as the record population, to which the adversary has sampling access.

To instantiate the MIA game, the challenger samples a set of $I$ users $\mathcal{I}_{\mathrm{train}} \subset [U]$ uniformly at random.  
For each $u \in \mathcal{I}_{\mathrm{train}}$, the challenger samples a dataset $\mathcal{D}_u \sim \pi_u^{n_u}$ and concatenates these to form the training set
\begin{equation*}
\mathcal{D}_{\mathrm{train}} = \bigcup_{u \in \mathcal{I}_{\mathrm{train}}} \mathcal{D}_u.
\end{equation*}
Here, we set $n_u = n$ for all $u$.  
Analogously, a validation dataset $\mathcal{D}_{\mathrm{val}}$ is constructed from users in $\mathcal{I}_{\mathrm{val}} \subset [U] \setminus \mathcal{I}_{\mathrm{train}}$.  
The remaining users are collected in
$\mathcal{I}_{\mathrm{test}} = [U] \setminus \left( \mathcal{I}_{\mathrm{train}} \cup \mathcal{I}_{\mathrm{val}} \right)$,
and a corresponding test dataset $\mathcal{D}_{\mathrm{test}}$ is created in the same way.
Finally, the challenger trains a model $f_{\theta}$, where $\theta = \mathcal{T}(\mathcal{D}_{\mathrm{train}})$.

The adversary has black-box access to the model $f_\theta$, i.e., it can obtain the forecast for arbitrary input queries but does not have access to the weights $\theta$.  
Furthermore, the adversary is assumed to know the training procedure, including the hyperparameters and model architecture of the target model.

We distinguish between record-level and user-level MIA.
In record-level MIA, the adversary goal is to devise an attack $\mathcal{A}$ to infer if a given data record $(X,Y)$ was included in the training set $\mathcal{D}_{\mathrm{train}}$ of $f_{\theta}$.
The privacy game instantiation is illustrated in Game~\ref{game:mia-game}.
Notably, the adversary does not have access to the weights $\theta$ but can only query the model $f_\theta$, and may use the record-population $\pi$  in the attack.



In user-level MIAs, formalized in Game~\ref{game:user-mia-game}, the adversary attempts to determine whether a collection of records belonging to a single user was included in the training set, thereby inferring the membership status of that user. 
We note that, to setup the attack, the adversary has access to the record population, $\pi$, rather than the user populations $\lbrace \pi_u \rbrace_{u\in[U]}$, i.e., it is unable to sample from a given user distributions.

\begin{game}[t]
\caption{Record-Level Membership Inference Game}
\label{game:mia-game}
\textbf{Input:}  Population distribution $\pi$, number of users in the population $U$, user population distributions $\lbrace \pi_u \rbrace_{u\in[U]}$, training algorithm $\mathcal{T}$, number of users for training $I$, dataset size $n$.\\
\textbf{Output:} Adversary’s guess $\tilde{b}$. 

\begin{algorithmic}[1]
\State Sample $I$ users into $\mathcal{I}_{\mathrm{train}}$ uniformly at random from $ [U]$
\State Create training dataset $\mathcal{D}_{\mathrm{train}}\leftarrow\cup_{u\in\mathcal{I}_{\mathrm{train}}} \mathcal{D}_u$, $\mathcal{D}_u \sim\pi_u^{n}$ 
\State Train model parameters: $\theta \leftarrow \mathcal{T}(\mathcal{D}_{\mathrm{train}})$
\State Sample member record $(X_1, Y_1) \sim \mathcal{D}_{\mathrm{train}}$
\State Sample non-member record $(X_0, Y_0) \sim \mathcal{D}_{\mathrm{test}}$
\State Sample random bit $b \sim \mathrm{Bernoulli}(0.5)$
\State Challenger sends $(\mathcal{T}, f_\theta, (X_b, Y_b))$ to the adversary
\State Adversary infers membership $\tilde{b} \leftarrow \mathcal{A}(\mathcal{T}, \pi,  f_\theta, (X_b, Y_b))$
\end{algorithmic}
\end{game}

\begin{game}[t]
\caption{User-Level Membership Inference Game}
\label{game:user-mia-game}

\textbf{Input:} Population distribution $\pi$, number of users in the population $U$, user population distributions $\lbrace \pi_u \rbrace_{u\in[U]}$, training algorithm $\mathcal{T}$, number of users for training $I$, dataset size $n$, challenge dataset size $n_{\mathrm{c}}$.  \\
\textbf{Output:} Adversary’s guess $\tilde{b}$

\begin{algorithmic}[1]
    \State Sample $I$ users into $\mathcal{I}_{\mathrm{train}}$ uniformly at random from $ [U]$
    \State Create training dataset $\mathcal{D}_{\mathrm{train}}\leftarrow\cup_{u\in\mathcal{I}_{\mathrm{train}}} \mathcal{D}_u$, $\mathcal{D}_u \sim\pi_u^{n}$ 
    \State Train model parameters: $\theta \leftarrow \mathcal{T}(\mathcal{D}_{\mathrm{train}})$
    \State Sample a member user $u_1 \sim \mathcal{I}_{\mathrm{train}}$
    \State Sample a non-member user $u_0 \sim \mathcal{I}_{\mathrm{test}}$
    \State Sample a random bit $b \sim \mathrm{Bernoulli}(0.5)$
    \State Sample a challenge dataset $\tilde{\mathcal{D}}_{u_b} \sim \pi_{u_b}^{n_{\mathrm{c}}}$ of size $n_{\mathrm{c}}\leq n$ 
    \State Challenger sends $(\mathcal{T}, f_\theta, \tilde{\mathcal{D}}_{u_b})$ to the adversary
    \State Adversary infers membership $\tilde{b} \leftarrow \mathcal{A}(\mathcal{T}, \pi, f_\theta,\tilde{\mathcal{D}}_{u_b})$
\end{algorithmic}
\end{game}

%% file: 05.tex
\section{Method}\label{sec:method}
We propose two attacks for time series forecasting models: a statistics-based and a classifier-based.  
The attacks are described from the perspective of record-level attacks, and their extension to user-level attacks is discussed in Section~\ref{sec:sample-user}.
Both attacks rely on shadow models that are trained on datasets created in the same way as the target dataset but using different user-splits. 
We denote the $i$th shadow model by $\phi_i$.

\subsection{Multi-Signal LiRA} \label{sec:multi-lira}
As discussed in Section~\ref{sec:background-signals}, a variety of statistical signals can be extracted from time series forecasts to serve as features for membership inference.  
Building on this, we propose a multi-signal variant of LiRA, inspired by prior work that uses likelihood scores from multiple augmentations of the same record to improve inference performance~\cite[Sec.~IV.C]{carlini2022first}.

Let $\mathcal{S}$ denote the set of available attack signals.  
For a given record $(X,Y)$, the model predicts $\hat{Y} = f_{\theta}(X)$, from which a signal is computed according to $s(Y, \hat{Y})$, $s \in \mathcal{S}$.  
Note that some signals in $\mathcal{S}$ may not make use of $\hat{Y}$, in which case it is simply ignored.

For the attack, we train $K$ shadow models to mimic the target model $f_\theta$.  
The adversary runs each record through both the target model $f_{\theta}$ and each shadow model $f_{\phi_i}$, $i \in [K]$, to yield $K\!+\!1$ predictions.  
Let $s(Y, f_{\phi_i}(X))$ denote the value of signal $s$ evaluated on the output of shadow model $i$.  
We compute the mean and standard deviation for each signal type over the shadow models.  
Shadow models are divided into in-models, which include $(X,Y)$ in their training data, and out-models, which do not. For each group, we compute the mean and standard deviation of the signal values, yielding $(\mu_v^s, \sigma_v^s)$ where $v \in \{\text{in}, \text{out}\}$ and $s \in \mathcal{S}$~\cite{carlini2022first}.

We model the signals for each group $v$ as multivariate Gaussian distributed, $\mathbf{s} \sim \mathcal{N}({\mu}_v, \Sigma_v)$, where ${\mu}_v = (\mu_v^s)_{s \in \mathcal{S}}$ and $\Sigma_v$ is the empirical covariance matrix of the signals in group $v$.  
A principled approach would be to estimate the full covariance matrix $\Sigma_v$, thereby capturing dependencies between signals.
However, stable estimation of the off-diagonal entries of $\Sigma_v$ typically requires a large number of shadow models; prior work on text models has used more than $400$ to achieve reliable estimates~\cite{rossi2025membership}.
Given the practical constraint that we can only train a limited number of shadow models, we approximate $\Sigma_v$ by a diagonal matrix containing only the variances $(\sigma_v^s)^2$ for each $s \in \mathcal{S}$.  

For the online attack, where shadow models may include the target data, the resulting likelihood-ratio score is  
\begin{equation}
\label{eq:mia_score}
\text{score}(X, Y; f_\theta) = 
\frac{p\!\left( \mathbf{s}(X,Y; f_\theta) \,\middle|\, 
\mathcal{N}\!\big({\mu}_{\mathrm{in}}, \Sigma_{\mathrm{in}}\big) \right)}
     {p\!\left( \mathbf{s}(X,Y; f_\theta) \,\middle|\, 
\mathcal{N}\!\big({\mu}_{\mathrm{out}}, \Sigma_{\mathrm{out}}\big) \right)},
\end{equation}
where $\mathbf{s}(X,Y; f_\theta) = \big(s(Y,f_\theta(X))\big)_{s \in \mathcal{S}}$ is the vector of signal values for record $(X,Y)$.

For the offline attack, the score is defined as the multivariate Gaussian tail probability under the out-model distribution:  
\begin{equation}
\label{eq:mia_score_offline}
\text{score}(X, Y; f_\theta) = 
\Pr\!\big(\mathbf{Z} \preceq \mathbf{s}(X,Y; f_\theta)\big),  \mathbf{Z} \sim \mathcal{N}\!\big({\mu}_{\mathrm{out}}, \Sigma_{\mathrm{out}}\big),
\end{equation}
where $\preceq$ denotes element-wise inequality.  
When $\Sigma_{\mathrm{out}}$ is diagonal, this reduces to the product of per-signal univariate tail probabilities.

\subsection{Deep Time Series Attack}

MIAs for time series forecasting typically rely on manually engineered statistical features extracted from model outputs, including our proposed attack in Section~\ref{sec:multi-lira}. 
These approaches require domain-specific feature selection and may fail to capture subtle patterns present in the raw temporal structure of predictions and their corresponding errors. To address this limitation, we propose the DTS Attack, a classifier-based approach that bypasses manual signal extraction. DTS directly learns to infer membership from model outputs by training a deep-learning classifier. 
The procedure is formalized in Algorithm~\ref{alg:dts}.

The attack is based on supervised training of a classifier that learns to distinguish members from non-members given the output of a forecasting model.  
To construct the training dataset, we assume the adversary has access to an auxiliary dataset $\mathcal{D}_{\mathrm{aux}}$.
In the online setting, $\mathcal{D}_{\mathrm{aux}}$ constitutes all of $\mathcal{D}$ whereas, in the offline setting, $\mathcal{D}_{\mathrm{aux}}$ contains no records from $\mathcal{D}_{\mathrm{train}}\cup\mathcal{D}_{\mathrm{val}} \cup \mathcal{D}_{\mathrm{test}}$.  
From $\mathcal{D}_{\mathrm{aux}}$, $K$ (potentially overlapping) datasets $\mathcal{D}_i$ are sampled and used to train shadow models $\{\phi_i\}_{i=1}^K$ (Step~1 in Algorithm~\ref{alg:dts}).

To construct the training dataset $\mathcal{D}_{\mathrm{DTS}}$, for each shadow model $\phi_i$, $i \in [K]$, we randomly sample a fraction $f$ of records from $\mathcal{D}_{\mathrm{aux}}$. Each sampled record $(X,Y)$ is evaluated by $\phi_i$ to produce the triplet $(Y, f_{\phi_i}(X), \ell_i)$, where $\ell_i = \mathbf{1}\lbrace (X,Y) \in \mathcal{D}_i\rbrace \in \lbrace 0,1 \rbrace$ indicates whether $(X,Y)$ was used to train $\phi_i$. These triplets are added into $\mathcal{D}_{\mathrm{DTS}}$, with each entry in $(\mathbb{R}^{M \times H} \times \mathbb{R}^{M \times H} \times \lbrace 0,1 \rbrace)$, yielding a dataset of size $K f |\mathcal{D}_{\mathrm{aux}}|$.

The membership classifier $\theta_{\mathrm{DTS}}$ is then trained on $\mathcal{D}_{\mathrm{DTS}}$ to map the $(Y,\hat{Y})$ pairs into a membership probability,
\[
\theta_{\mathrm{DTS}}: \left( \mathbb{R}^{M\times H} \times \mathbb{R}^{M\times H} \right) \longrightarrow [0,1],
\]
as shown in Steps~2–3 of Algorithm~\ref{alg:dts}.  
Through this training process, the classifier learns to distinguish between training and non-training records by leveraging temporal dependencies and statistical regularities in the joint distribution of model predictions and true outcomes.

After training, $\theta_{\mathrm{DTS}}$ can be used to attack the target model $f_\theta$.  
Given a target record $(X, Y)$, the target model prediction $\hat{Y} = f_\theta(X)$ is computed, after which $(Y, \hat{Y})$ is passed to $\theta_{\mathrm{DTS}}$ to obtain the membership inference score
\begin{equation}
\mathrm{score}\big(X, Y; f_\theta\big) 
= \theta_{\mathrm{DTS}}\big(Y, \hat{Y}\big),
\end{equation}
which represents the estimated probability that $(X, Y)$ was part of the training set of the target model.  
The attacker then compares this probability to a decision threshold to determine membership.

\begin{algorithm}[t]
\caption{Deep Time Series (DTS) MIA }
\label{alg:dts}
\begin{algorithmic}[1]

\Statex \textbf{Input:} Target model $f_\theta$, auxiliary dataset $\mathcal{D}_{\mathrm{aux}}$, sampling fraction $f$, training algorithm $\mathcal{T}$, number of shadow models $K$, target data $(X,Y)$
\Statex \textbf{Output:} Membership score for the target data $(X,Y)$

\Statex \textbf{Step 1: Train shadow models}
\For{$i = 1 \dots K$}
    \State Sample a training subset $\mathcal{D}_i$ from $ \mathcal{D}_{\mathrm{aux}}$
    \State Train shadow model $\phi_i \leftarrow \mathcal{T}(\mathcal{D}_i)$
\EndFor

\Statex \textbf{Step 2: Construct DTS training dataset $\mathcal{D}_{\mathrm{DTS}}$}
\State $\mathcal{D}_{\mathrm{DTS}} = \emptyset$
\For{$i=1 \dots K$} 
    \State Sample a fraction $f$ from $\mathcal{D}_{\mathrm{aux}}$ into $\tilde{\mathcal{D}}$ 
    \For{each $(\tilde{X}, \tilde{Y}) \in \tilde{\mathcal{D}}$} 
        \State Compute prediction $\hat{Y} = f_{\phi_i}(\tilde{X})$
        \State Assign $\ell_i = 1$ if $\tilde{X} \in \mathcal{D}_i$, otherwise $\ell_i = 0$
        \State Add tuple $(\tilde{Y}, \hat{Y}, \ell_i)$ to $\mathcal{D}_{\mathrm{DTS}}$
    \EndFor
\EndFor

\Statex \textbf{Step 3: Train the classifier}
\State Train $\theta_{\mathrm{DTS}}$ using $\mathcal{D}_{\mathrm{DTS}}$

\Statex \textbf{Step 4: Score the target record}
    \State Compute target prediction $\hat{Y} = f_\theta(X)$
    \State Obtain score: $
    \mathrm{score}\big(X, Y; f_\theta\big) 
    = \theta_{\mathrm{DTS}}\big(Y, \hat{Y}\big)
    $
\end{algorithmic}
\end{algorithm}

\subsection{Record-Level to User-Level MIA}
\label{sec:sample-user}

The two attacks described in the previous sections output \textit{record-level} membership inference scores, i.e., the estimated probability that a given record from the audit dataset was included in the target model’s training set.  
Extending this formulation to the user level introduces an additional challenge, as each user typically contributes multiple records.  
User-level membership decisions must therefore take into account several record-level scores from the same user, which may be combined in various ways.

Several combination strategies can be used, including simple majority voting, averaging or taking the median of record-level scores, or emphasizing extreme outliers. 
In this work, we adopt a likelihood-based approach, computing the product of the record-level membership probabilities for a given user. Under the simplifying assumption that record-level predictions are independent, this product corresponds to the overall likelihood that the user’s dataset was included in training.

Formally, a user $u \in \mathcal{I}_{\mathrm{train}}$ contributes $\mathcal{D}_u = \{ (X_{u,1}, Y_{u,1}), \dots, (X_{u,n}, Y_{u,n}) \}$, consisting of $n$ training records.
During an attack on user $u$, the adversary observes a collection of $n_{\mathrm{c}}$ records $\tilde{\mathcal{D}}_u \sim \pi_u^{n_{\mathrm{c}}}$ whose elements may coincide with those in $\mathcal{D}$.
The user-level membership score for user $u$ is then given as
\begin{align}
    \text{score}_{\mathrm{user}}\!\left(\tilde{\mathcal{D}}_u; f_\theta\right)
    = \prod_{(X,Y) \in \tilde{\mathcal{D}}_u} 
      \text{score}\left(X, Y ;f_\theta\right),
    \label{eq:indiv_mia}
\end{align}
where $\text{score}\!\left(X, Y ;f_\theta\right)$ is the 
record-level membership probability for a single forecasting record $(X,Y)$.

%% file: 06.tex
\section{Experimental Setup}

\subsection{Data Setup and Preprocessing}
\label{sec:datasets}
We evaluate our attacks on two real-world time series datasets from privacy-sensitive domains, each requiring domain-specific preprocessing. The first consists of electroencephalogram (EEG) signals which are biometric by nature and highly distinctive across individuals, making them inherently identifiable~\cite{jalaly2020eeg}. The second, electricity load data, while non-biometric, can still reveal extensive private information: prior work has shown that fine-grained consumption traces can be exploited to infer household activities, device usage, occupancy patterns, and even socio-economic or employment status~\cite{voyez2022grid}.
We describe both datasets in detail below.

\subsubsection{The Temple University Hospital EEG Data Corpus (TUH-EEG v2.0.1)}
The TUH-EEG data corpus is a clinical dataset containing $26846$ EEG recordings from around $1500$ individuals, collected between 2002 and 2017~\cite{obeid2016tuheeg}. The majority of the EEG recordings contains $31$ channels, however, some have as few as $20$. 
Most signals were sampled at $250$ Hz ($87\%$), with smaller portions recorded at $256$ Hz ($8.3\%$), $400$ Hz ($3.7\%$), and $512$ Hz ($1\%$).

To ensure data quality, we selected a subset of $100$ recordings from the first $300$ individuals with recordings sampled at $250$ Hz. 
We removed the initial $60$ seconds of each recording which often contain artifacts such as calibration waves, and excluded sessions with persistent noise or flat channels.

Following prior work~\cite{koren2024membership}, we selected $M=3$ frontal EEG leads (FP1, FP2, F3) and truncated recordings to $15000$ time steps. 
Records were generated using a sliding window with a lookback of $L=100$, using a stride of $1$, and a prediction horizon $H=20$, see Section~\ref{sec:forecasting}.

All time series were normalized using interquartile range (IQR) scaling via the \texttt{RobustScaler} from scikit-learn~\cite{robustscaler}, which we found to perform best empirically due to its robustness against outliers and inter-subject variability.

\subsubsection{Electricity Load Diagrams (ELD) Dataset}
The ELD dataset~\cite{electricity} is a widely used benchmark for time series forecasting. 
It is based on recordings of electricity consumption from $370$ Portuguese households over a four-year period (2011–2014), with a sampling resolution of $15$ minutes. Measurements are expressed in kilowatts. Since not all households were monitored throughout the entire period, some recordings contain zero-padding to make all time series equal length.

To prepare the data, we removed the zero-padding and aggregated the time resolution to $1$ hour by summing every four consecutive measurements. Households with fewer than $15000$ valid time steps or with average usage below $200$ kW/h or above $2000$ kW/h were excluded to avoid unreliable signals and extreme outliers. We then randomly selected $100$ households for our experiments. Each series was truncated to $15000$ time steps and normalized using IQR scaling, following the same procedure used for EEG data.

The dataset constitutes a univariate time series, i.e., $M=1$, and records were generated via a sliding window, using a stride of $1$, with a lookback of $L=100$ and a prediction horizon of $H=20$, as described in Section~\ref{sec:forecasting}. 

\subsection{Data Split}
After preprocessing, each dataset contains time series data from $I=100$ individuals. 
For each individual, the window operation results in $14881$ data points per individual.
We adopt an individual-based partitioning strategy to ensure that no individual's data appears in both the train and test datasets. 
In particular, we divide the $100$ individuals into a training dataset $\mathcal{D}_{\mathrm{train}}$, validation dataset $\mathcal{D}_{\mathrm{val}}$, and test dataset $\mathcal{D}_{\mathrm{test}}$ of $20$ individuals each. 
We emphasize that we make use of the validation dataset for early-stopping to limit the amount of overfitting, hence, making the attack setting more difficult.
The remaining $40$ individuals constitute the auxiliary dataset $\mathcal{D}_{\mathrm{aux}}$ and is used by the adversary to train shadow models.
During the attack, we make use of $\mathcal{D}_{\mathrm{train}} \cup \mathcal{D}_{\mathrm{test}}$ to sample records to infer the membership of.
In the online setting, each shadow model is trained on a randomly selected $50\%$ subset of individuals from $\mathcal{D}_{\mathrm{train}} \cup \mathcal{D}_{\mathrm{test}}$ instead of sampling from $\mathcal{D}_{\mathrm{aux}}$ as in the offline setting.
For user-level attacks, we let $\tilde{\mathcal{D}}_u = \mathcal{D}_u$, i.e., the adversary observes all user data records during the attack. 
We consider partial observations in the ablations.

\subsection{Target Model Architectures}
We assess membership inference risks in time series forecasting using two representative deep learning architectures: Long Short-Term Memory (LSTM)~\cite{hochreiter1997lstm} and Neural Hierarchical Interpolation for Time Series (N-HiTS)~\cite{olivares2023nhits}.
N-HiTS is a hierarchical, multi-resolution forecasting model that extends N-BEATS~\cite{oreshkin2020nbeats} with multi-rate input sampling and multi-scale interpolation, achieving higher accuracy and efficiency than prior methods in long-horizon forecasting while supporting multivariate and exogenous inputs.

By selecting these two models, we cover both an established baseline and a state-of-the-art approach to time series modeling. Both are implemented in PyTorch and configured for multivariate sequences with varying input lengths and prediction horizons. Their specific configurations and hyperparameters are detailed below.

\subsubsection{LSTM}
The LSTM model is a standard multivariate sequence-to-sequence model. It consists of two unidirectional LSTM layers, each with $64$ hidden units. The model accepts sequences of length $L$ with $M$ variables as input (i.e., $X \in \mathbb{R}^{M \times L}$), where the number of input features is set to $M$. The final hidden state from the second LSTM layer is passed to a fully connected layer with $H \times M$ output neurons, representing the forecast for the next $H$ time steps.

\subsubsection{N-HiTS}
For the N-HiTS model, we build on the official implementation~\cite{nhits_implementation}, applying minimal modifications to fit our experimental setup. Specifically, we remove dependencies on exogenous variables and adjust input dimension handling for compatibility with our preprocessed datasets.
The architecture consists of three stacks, each containing a single block with two fully connected layers of $512$ units. Each block employs the \texttt{IdentityBasis} function to directly map outputs into backcast and forecast components, without explicitly modeling trends or seasonality. We adopt the default hyperparameters recommended by the original authors: Xavier normal initialization~\cite{glorot2010understanding}, no weight sharing between blocks, no dropout or batch normalization, and ReLU activations. Linear interpolation is used as the pooling method across resolution levels.

\subsection{Training Protocol}
We trained the target models using the Adam optimizer with a default learning rate of $0.001$ and the MAE loss as the optimization criterion. Training was performed for a maximum of $50$ epochs with early stopping based on validation loss, using a patience of $3$ epochs. A batch size of $1024$ was used for all training runs.

\subsection{Target Model Evaluation}
To evaluate the performance of our target models, we report four forecasting metrics: MSE, MAE, SMAPE, and normalized deviation (ND)~\cite{oreshkin2020nbeats}. MSE and MAE are standard point-wise metrics, while SMAPE and ND measure errors relative to the magnitude of the true values. Given our individual-level data partitioning and the substantial variance across subjects, these scale-sensitive metrics provide more informative indicators of overfitting and generalization.
Formally, for a prediction horizon $H$ and $M$ target variables, SMAPE and ND are defined as:
\begin{equation} \label{eq:smape}
\text{SMAPE}(Y, \hat{Y}) = \frac{1}{HM} \sum_{i=1}^H \sum_{j=1}^M
\frac{\lvert Y_{i,j} - \hat{Y}_{i,j} \rvert}{\lvert Y_{i,j} \rvert + \lvert \hat{Y}_{i,j} \rvert}
\end{equation}
and
\begin{equation}
\text{ND}(Y, \hat{Y}) =
\frac{\sum_{i=1}^H \sum_{j=1}^M \lvert Y_{i,j} - \hat{Y}_{i,j} \rvert}
{\sum_{i=1}^H \sum_{j=1}^M \lvert Y_{i,j} \rvert}.
\label{eq:nd}
\end{equation}

\subsection{Attacks Implementation}

We evaluate attack performance using three state-of-the-art baselines, and our proposed attacks from Section~\ref{sec:method}, employing $64$ shadow models for all configurations. Their specific instantiations in our experimental setup are described below.

All attacks leverage the signals introduced in Section~\ref{sec:background-signals}. For some attacks, the underlying signal must be unbounded; to accommodate this, we define a rescaled version of SMAPE:
\begin{equation}\label{eq:rsmape}
\text{rSMAPE} = \log \left( \frac{\text{SMAPE}(Y, \hat{Y})}{1 - \text{SMAPE}(Y, \hat{Y})} \right).
\end{equation}

\subsubsection{Baseline 1 (LiRA)}
We conduct multiple instances of the LiRA attack~\cite{carlini2022first}, substituting the rescaled logit from the original formulation with each of the signals introduced in Section~\ref{sec:background-signals} and in~\eqref{eq:rsmape}.
These variants are evaluated across all datasets and target models in both online and offline settings. Variance for the in- and out-distributions is estimated following the per-example variance method~\cite[Sec.~VI.B]{carlini2022first}.

\subsubsection{Baseline 2 (RMIA)}
RMIA~\cite{zarifzadeh2024lowcost} compares the likelihood ratio between a model trained with the target record and one where it is replaced by a random population record, using its tail probability as the test score.
The attack instantiations use six signals: MSE, MAE, SMAPE, Trend, Seasonality, and TS2Vec, evaluated in both online and offline settings across all datasets and target models. 
Since $\text{rSMAPE}$ is unbounded, it is omitted from RMIA.
For RMIA, which operates on probabilities, we apply a monotonically decreasing mapping  
\begin{equation*}
    g(s(X,Y; f_\theta)) = \frac{1}{1 + s(X,Y; f_\theta)}
\end{equation*}
to convert signal values into probability scores. RMIA also requires sampling from the population to estimate the average shadow-model output; for this, we draw records from the dataset excluding all train, test, and validation instances.
We set $\gamma = 1.0$ and $\alpha = 1/3$ for the offline setting~\cite[App.~B.2.2]{zarifzadeh2024lowcost}.

\subsubsection{Baseline 3 (Ensemble Attack)}
Shachor et al.~\cite{sachor2024improved} introduce an Ensemble attack, which assumes the adversary has access to a dataset of in-members (target training data points) along with non-members. The method trains multiple small, specialized classifiers on disjoint subsets of data, each balanced with 50\% members and non-members, to improve membership inference. For each subset, several  classifiers are trained on target model outputs, or derived features, evaluated, and the best one retained; repeating this across many subsets yields an ensemble whose score for a point is the fraction of models predicting it as a member. This approach is adopted by Koren et al.~\cite{koren2024membership}, who are the first to study time series forecasting in this context. 
In our work, we use the same attack parameters as~\cite{koren2024membership}, apply mean aggregation within the ensemble as in~\cite{sachor2024improved}, and select the best model based on ROC-AUC performance.
Specifically, we set the number of complete executions of the attack on different random records to five, the number of train/test repetitions per subset pair to three, the subset size to $50$, and the number of fixed member–non-member subset combinations per instance to nine.

Next, we consider the attacks proposed in Section~\ref{sec:method}.

\subsubsection{Multi-Signal LiRA}
We applied the Multi-Signal LiRA attack (Section~\ref{sec:multi-lira}) to all datasets and target models under both online and offline evaluation settings. The signal vector comprises MSE, MAE, \rSMAPEtext, Trend, Seasonality, and TS2Vec. We assumed a diagonal covariance matrix and estimate each entry using the per-example variance method of~\cite[Sec.~VI.B]{carlini2022first}.\footnote{We also experimented with using a full covariance matrix, but the limited number of shadow models in our setting prevented stable estimation. While~\cite{rossi2025membership} apply a similar idea to LLMs, stabilizing covariance estimation via shrinkage with over 400 shadow models, we intentionally restrict our setup to a scale that remains feasible across the multiple datasets, models, and attack variants considered in this study.}

\subsubsection{Deep Time Series Attack}
The DTS attack is applied to all datasets and target models under both online and offline evaluation settings. For each configuration, we construct $\mathcal{D}_{\mathrm{DTS}}$ by sampling $f=10\%$ from the data set governing the creation of shadow datasets. We select a small $f$ to demonstrate that the method performs strongly even at low values; the performance is expected to improve as $f$ increases.
Each DTS classifier is trained with the Adam optimizer, using early stopping (patience $=3$) and a maximum of 64 epochs.

We evaluate two classifier architectures: a simple LSTM-based model and InceptionTime~\cite{fawaz2020inceptiontime}, a convolution-based architecture that, along with its variants, is considered state-of-the-art for time series classification on the UCR archive~\cite{ismail2025liteInception} which is a collection of multiple time series datasets~\cite{dau2019ucr}.
In contrast to prior work~\cite{fawaz2020inceptiontime}, which relies on an ensemble of predictors, we simplify the approach by employing a single one.

%% file: 07.tex
\section{Experimental Results}
\label{sec:experimental_results}

\subsection{Target Models}

Table~\ref{tb:target_model} reports target model performance on the TUH-EEG and ELD datasets across training, validation, and test splits, with results averaged over five runs to account for variability from random initialization and dataset partitioning. 
The randomization includes both the target model’s data split and the user selection; in each run, a new subset of users is sampled ($100$ out of $300$ for TUH-EEG and $100$ out of $370$ for ELD).
The similar performance gaps between splits for both architectures indicate that early stopping effectively prevents overfitting. Across all metrics (MSE, MAE, SMAPE, and ND), N-HiTS outperforms LSTM. 
On TUH-EEG, the largest gains for N-HiTS are observed in MSE (a reduction from $4.15 \times 10^{-9}$ to $8.43 \times 10^{-10}$) and ND (from $0.319$ to $0.223$) on the test set. On ELD, improvements are most pronounced in MSE (from $1.85 \times 10^4$ to $1.37 \times 10^{4}$) and SMAPE (from $0.0754$ to $0.0645$). 
MSE and MAE show higher variance across runs, while the scale-invariant SMAPE and ND remain more stable, making them better suited for cross-dataset comparison.

\input{Results_includes/target_model_table}

\subsection{Attack Results}
\label{sec:attack-results}
The performance of the different attacks is presented in Table~\ref{tab:eeg_results_summary} for the TUH-EEG dataset and Table~\ref{tab:eld_results_summary} for the ELD dataset, and in Table~\ref{tab:rocauc}. 
We follow earlier works~\cite{carlini2022first, zarifzadeh2024lowcost, lassila2025practical} and report the true positive rate (TPR) at a given FPR (Table~\ref{tab:eeg_results_summary} and Table~\ref{tab:eld_results_summary}) and the ROC-AUC score in Table~\ref{tab:rocauc}.
For the complete ROC curves, we refer to Appendix~\ref{sec:appendix_roc}.
In particular, for record-level attacks, we report TPR at an FPR of $0.1\%$ and $0.01\%$ while, for user-level attacks, we report the TPR at $0\%$. All user-level results correspond to an access ratio of $g =1$. 
All results are averaged over five independent runs, with both the mean and standard deviation reported. 
For each attack type, we illustrate the strongest configuration with boldface and the overall best attack for each reported metric is highlighted in \colorbox{myblue}{\phantom{XX}}  and \colorbox{myorange}{\phantom{XX}} for the online and offline setting, respectively.

\input{Results_includes/attacks_perfromance_eeg_table}

\input{Results_includes/attacks_perfromance_eld_table}

\input{Results_includes/ROCs_AUC_table}

\subsubsection{Data Sensitivity}

Attacks are consistently more effective on TUH-EEG than on ELD, a difference likely driven by several factors. First, our EEG experiments use three input channels versus a single channel for ELD, so with the lookback and horizon fixed, more input information is available to the MIA. Second, as discussed in Section~\ref{sec:datasets}, EEG signals are highly individual-specific, whereas electricity usage, though also carrying identifiable patterns, is less distinctive due to overlapping daily routines and shared household behaviors. 
Finally, EEG data has much finer temporal resolution ($4$ms) compared to ELD ($1$h), providing attackers with far more granular input. Together, these factors help explain the stronger attack performance observed on EEG-trained models, as forecasting models naturally adapt to stable user-specific temporal structure when it is present.

\subsubsection{Comparing Attack Strategies}
The Ensemble attack~\cite{sachor2024improved} performs poorly under the stringent metrics considered, ranking near the bottom across datasets and settings. Its limited per-model training data and lack of adaptability likely explain its weak performance.

RMIA~\cite{zarifzadeh2024lowcost} stands out in several scenarios. It achieves the highest user-level performance on TUH-EEG across both models and is the strongest method in the online setting for N-HiTS on ELD. At the record level, RMIA dominates offline attacks for N-HiTS on TUH-EEG and for all models on ELD. However, in the online record-level setting, RMIA is frequently outperformed by LiRA, Multi-Signal LiRA, and DTS.
Considering ROC-AUC in Table~\ref{tab:rocauc}, RMIA emerges as the clear winner in the offline setting.

For both datasets, LiRA performs competitively at the user level, sometimes matching the best results, but generally lags behind the strongest methods at the record-level, especially in the offline setting. 
In the online setting, TS2Vec often yields LiRA’s best record-level results, while \rSMAPEtext\ tends to perform better offline. No single signal dominates across all cases, and Trend and SMAPE are typically the weakest. 
A notable example is TUH-EEG with N-HiTS in the online setting, where all signals except Trend achieve a perfect $100\%$ user-level TPR at $0\%$ FPR across all seeds. 
On ELD, LiRA’s record-level scores are often surpassed by DTS or RMIA, but it remains competitive at the user level.

Our proposed Multi-Signal LiRA improves over the best single-signal LiRA in many online record-level settings but rarely outperforms the top-performing non-LiRA methods. 
However, with respect to ROC-AUC, Table~\ref{tab:rocauc}, it achieves the best performance against an LSTM target model.
In the offline setting, it can underperform compared to the strongest individual signals, such as \rSMAPEtext\ and TS2Vec. We anticipate that a fully multivariate LiRA, able to model correlations between signals, could further improve performance, as the signals are anticipated to correlate. 
However, because our current implementation models the covariance matrix as diagonal, due to the limited number of shadow models, these correlations cannot be fully exploited.

The DTS attack offers a strong deep learning–based alternative, achieving some of the highest online record-level TPRs across both datasets, particularly with the InceptionTime architecture. For example, on TUH-EEG with N-HiTS, DTS reaches $23.32\%$ TPR at $0.1\%$ FPR in the online setting, outperforming all other methods by a substantial margin. 
With respect to ROC-AUC, it achieves the strongest performance against the N-HiTS architecture, achieving up to $0.884$.
DTS is less competitive in the offline setting, where RMIA often dominates, but still achieves respectable results. Its performance depends heavily on the membership inference classifier architecture, with InceptionTime often outperforming LSTM. 

In Appendix~\ref{sec:DP-SGD}, we present results on the attack performance when the forecast models are trained using differential privacy.

\subsubsection{Offline vs Online Attacks}

Across all evaluations, online attacks consistently outperformed their offline counterparts, with especially large gains at the user level. This gap is substantially larger than what has been reported for other data modalities~\cite{carlini2022first,zarifzadeh2024lowcost, lassila2025practical}, 
suggesting that membership inference in time series forecasting may be particularly sensitive to distributional shifts. A likely factor is our individual-based partitioning. With only 100 diverse users, temporal and user-specific patterns can introduce significant distribution differences between the train/test and auxiliary sets, in contrast to the record-level splits commonly used in prior work. We further explore this hypothesis in Section~\ref{sec:ablation}.

For both datasets, the strongest record-level results in the online setting are obtained by LiRA and DTS, while RMIA and LiRA dominate at the user level. The magnitude of the online–offline gap can be striking. For example, on TUH-EEG with N-HiTS at $\text{FPR} = 0.01\%$, LiRA achieves a TPR of $2.64\%$ online compared to only $0.13\%$ offline. The difference is even more pronounced for DTS, which reaches $15.81\%$ online compared to $0.67\%$ offline.

\subsubsection{User vs Record-Level Attacks}

Our results show a consistent gap between record-level and user-level MIAs; it is substantially easier to infer membership status of a user than an individual record.  
This is in line with previous works~\cite{mireshghallah2022quantifying, meeus2024did} that argue for user-level attacks, that aggregate predictions across multiple records, providing stronger signals for membership inference.
Crucially, this discrepancy highlights a potential pitfall in evaluation since only considering record-level metrics may significantly underestimate the potential privacy risk. 
For instance, in the online setting against the N-HiTS model on the TUH-EEG dataset, LiRA achieves just 7.41\% TPR at 0.1\% FPR at the record-level, but 100\% TPR at 0\% FPR at the user level, indicating a far more severe vulnerability when considering membership on user level.

Another observation is the large variance observed in some user-level results. As shown in Section~\ref{sec:ablation}, this variance decreases with an increasing number of individuals, suggesting improved stability in larger populations. It is worth noting that user-level performance is reported at 0\% FPR, a statistically sensitive threshold. Because it allows no false positives, even minor fluctuations in model predictions can lead to significant changes in the measured TPR. Despite this instability, we report results at 0\% FPR to highlight that some attacks achieve perfect (100\%) user-level detection under this strict setting.

\subsection{Ablation Study}
\label{sec:ablation}

We conducted three ablation studies to examine how forecasting horizon, population size, and access ratio influence attack performance. The first two studies used an N-HiTS target model on the ELD dataset, while the third was performed on the EEG dataset, where user-level attacks achieve their strongest performance. Across all studies, we evaluated three representative online attacks: LiRA with TS2Vec, Multi-Signal LiRA, and DTS with InceptionTime.

\subsubsection{Impact of Forecasting Horizon}
\begin{figure*}[t!]
    \centering
    \resizebox{0.7\linewidth}{!}{%
    \input{Results_includes/fig_horizon}
    }
    \caption{Effect of forecasting horizon on record-level (left and mid) and user-level (right) TPRs for three attack configurations against the N-HiTS model on the ELD dataset.}
  \label{fig:horizon_ablation}
\end{figure*}
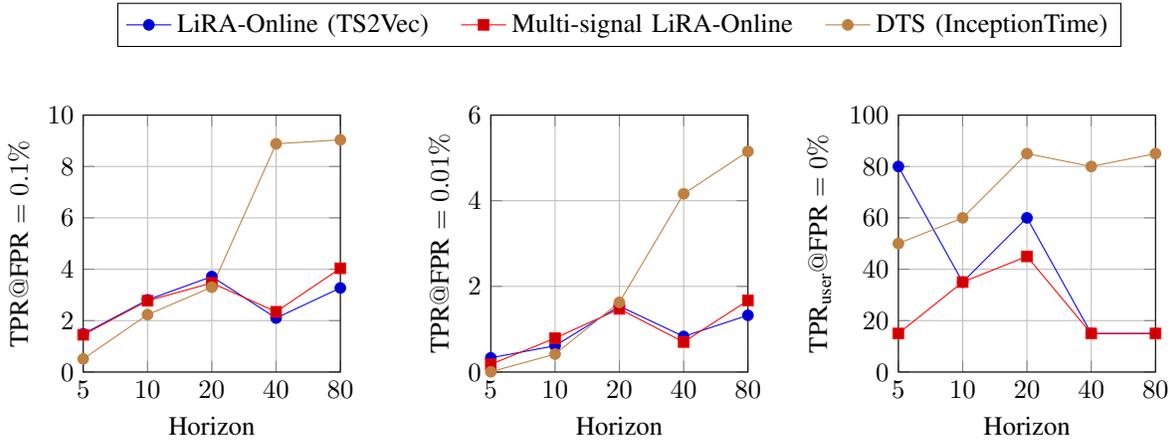

Figure~\ref{fig:horizon_ablation} reports record-level and user-level TPRs at fixed FPRs for varying forecasting horizons $H\in \lbrace 5,10,20,40,80\rbrace$. Longer horizons generally yield higher record-level TPRs, indicating increased information leakage with larger prediction windows. This trend is consistent across all attacks. At the user level, DTS continues to improve with longer horizons, whereas LiRA-based attacks peak at around horizon 20 before declining, possibly due to signal degradation or shifts in the underlying data distribution.
An increase in attack performance with sequence length has previously been documented in the context of LLMs~\cite{shi2023detecting, hayes2025strong}.

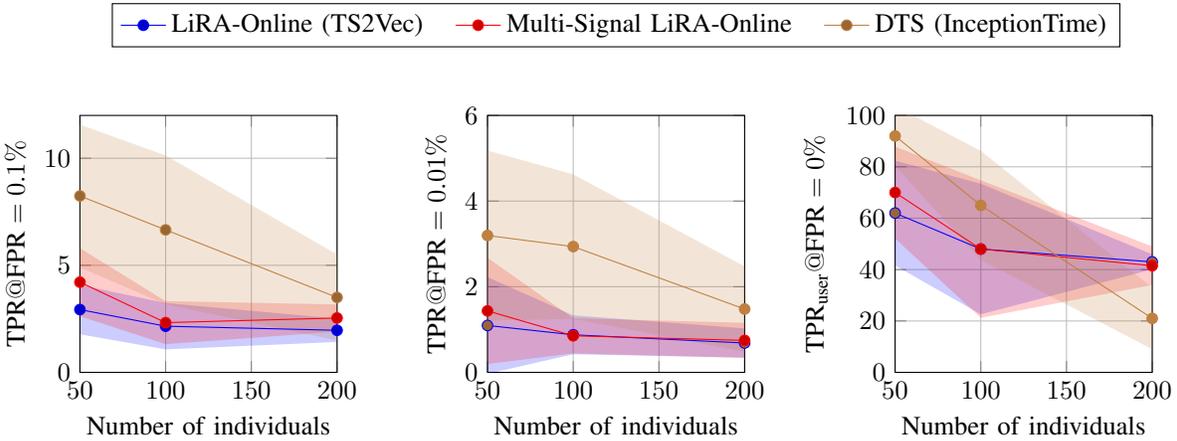
\begin{figure*}[t!]
    \centering
    \resizebox{0.7\linewidth}{!}{%
    \input{Results_includes/fig_individuals}
    }
    \caption{Effect of varying number of individuals on record-level (left and mid) and user-level TPRs (right) for LiRA-Online, Multi-Signal LiRA-Online, and DTS with InceptionTime; plotted with the confidence interval over five runs.}
  \label{fig:individuals_ablation}
\end{figure*}

\begin{figure}[t!]
    \centering
    \resizebox{0.55\columnwidth}{!}{%
    \input{Results_includes/fig_access_ratio_EEG}
    }
    \caption{Effect of observed dataset size on user-level MIA for LiRA-Online, Multi-Signal LiRA-Online, and DTS-Online.}
  \label{fig:access_ablation}
\end{figure}

\subsubsection{Impact of Number of Individuals}
 To examine the effect of population size, we varied the total number of users within $I\in\lbrace 50, 100, 200\rbrace$, keeping all other parameters fixed. Each configuration was evaluated over five random seeds to account for variation due to data splits. These totals correspond to $\mathcal{D}_{\mathrm{train}} \cup \mathcal{D}_{\mathrm{test}}$ with $20$, $40$, and $80$ individuals, respectively. 
 As shown in Figure~\ref{fig:individuals_ablation}, both record-level and user-level TPRs decrease as the number of individuals increases, suggesting that larger populations reduce vulnerability to MIAs under individual-based partitioning.

 \subsubsection{Impact of Partially Observed User Data}

We study the effect of the adversary’s observation size by letting 
$\tilde{\mathcal{D}}_u$ consist of subsets of $\mathcal{D}_u$ and varying
$n_{\mathrm{c}}$ as $n_{\mathrm{c}} \in \{0, 0.05n, 0.10n, \ldots, n\}$.
Here, $n_{\mathrm{c}} = 1$ reflects record-level inference with no user-level 
context, whereas $n_{\mathrm{c}} = n$ corresponds to the full-observation setting 
analyzed. Both of these settings are studied in Section~\ref{sec:attack-results}.

In Figure~\ref{fig:access_ablation}, we report the user-level performance,
averaged over five independent random subsets using the same target and shadow models.
As expected, attack performance increases with $n_{\mathrm{c}}$; more surprisingly, though, strong results emerge even at small fractions of a user's records, indicating a high susceptibility to user-level membership inference.

%% file: Results_includes/target_model_table.tex
\begin{table*}[t]
\centering
\begin{tabular}{l R{1.3cm} R{1.3cm} p{3.5cm} p{3.5cm} p{3.5cm} }
 \toprule
Datasets            & Models                  & Metrics & Train & Validation & Test \\ \hline
\multirow{8}{*}{TUH-EEG} & \multirow{4}{*}{LSTM}   
& MSE     
  & $\num{6.68e-10}$ {\scriptsize$\pm\num{4.46e-10}$} 
  & $\num{4.42e-9}$ {\scriptsize$\pm\num{4.87e-9}$} 
  & $\num{4.15e-9}$ {\scriptsize$\pm\num{2.20e-9}$} 
  \\
  &  
& MAE
  & $\num{1.28e-5}$ {\scriptsize$\pm\num{2.77e-6}$} 
  & $\num{2.05e-5}$ {\scriptsize$\pm\num{5.29e-6}$} 
  & $\num{2.30e-5}$ {\scriptsize$\pm\num{4.45e-6}$}  
\\

& &  SMAPE   
  & $\num{0.247}$ {\scriptsize$\pm\num{0.0285}$} 
  & $\num{0.271}$ {\scriptsize$\pm\num{0.0186}$} 
  & $\num{0.285}$ {\scriptsize$\pm\num{0.0111}$}     
\\
&& ND      
  & $\num{0.198}$ {\scriptsize$\pm\num{0.0323}$} 
  & $\num{0.296}$ {\scriptsize$\pm\num{0.0590}$} 
  & $\num{0.319}$ {\scriptsize$\pm\num{0.0063}$}     
\\ \cline{2-6} 
                   
& \multirow{4}{*}{N-HiTS} 
& MSE    
 & $\num{2.60e-10}$ {\scriptsize$\pm\num{1.47e-10}$} 
  & $\num{6.97e-10}$ {\scriptsize$\pm\num{3.04e-10}$} 
  & $\num{8.43e-10}$ {\scriptsize$\pm\num{5.10e-10}$} 
\\
&& MAE
& $\num{8.92e-6}$ {\scriptsize$\pm\num{1.43e-6}$} 
  & $\num{1.49e-5}$ {\scriptsize$\pm\num{2.45e-6}$} 
  & $\num{1.60e-5}$ {\scriptsize$\pm\num{2.46e-6}$}   
\\
&& SMAPE     
& $\num{0.208}$ {\scriptsize$\pm\num{0.0301}$} 
  & $\num{0.258}$ {\scriptsize$\pm\num{0.0195}$} 
  & $\num{0.267}$ {\scriptsize$\pm\num{0.0048}$}     
  \\
&& ND
  & $\num{0.139}$ {\scriptsize$\pm\num{0.0196}$} 
  & $\num{0.216}$ {\scriptsize$\pm\num{0.0294}$} 
  & $\num{0.223}$ {\scriptsize$\pm\num{0.0201}$}
  \\ \hline
\multirow{8}{*}{ELD} & \multirow{4}{*}{LSTM}   
& MSE    
& $\num{1.47e4}$ {\scriptsize$\pm\num{6.00e3}$} 
  & $\num{2.77e4}$ {\scriptsize$\pm\num{1.90e4}$} 
  & $\num{1.85e4}$ {\scriptsize$\pm\num{6.65e3}$}
  \\
&& MAE
 & $\num{65.8}$ {\scriptsize$\pm\num{12.9}$} 
  & $\num{88.0}$ {\scriptsize$\pm\num{25.0}$} 
  & $\num{73.8}$ {\scriptsize$\pm\num{6.9}$}
  \\
&& SMAPE  
& $\num{0.0607}$ {\scriptsize$\pm\num{0.0178}$} 
  & $\num{0.0668}$ {\scriptsize$\pm\num{0.0226}$} 
  & $\num{0.0754}$ {\scriptsize$\pm\num{0.0056}$}
  \\
&& ND
& $\num{0.0984}$ {\scriptsize$\pm\num{0.0177}$} 
  & $\num{0.116}$ {\scriptsize$\pm\num{0.0315}$} 
  & $\num{0.123}$ {\scriptsize$\pm\num{0.0089}$}
  \\ \cline{2-6}

& \multirow{4}{*}{N-HiTS} 
& MSE
 & $\num{7.92e3}$ {\scriptsize$\pm\num{2.80e3}$} 
  & $\num{1.95e4}$ {\scriptsize$\pm\num{1.29e4}$} 
  & $\num{1.37e4}$ {\scriptsize$\pm\num{3.45e3}$}    
\\
&& MAE
 & $\num{49.1}$ {\scriptsize$\pm\num{9.6}$} 
  & $\num{69.5}$ {\scriptsize$\pm\num{19.6}$} 
  & $\num{61.1}$ {\scriptsize$\pm\num{10.9}$}     
\\
&& SMAPE
  & $\num{0.0471}$ {\scriptsize$\pm\num{0.0164}$} 
  & $\num{0.0556}$ {\scriptsize$\pm\num{0.0250}$} 
  & $\num{0.0645}$ {\scriptsize$\pm\num{0.0088}$}     
\\
&& ND
  & $\num{0.0731}$ {\scriptsize$\pm\num{0.0126}$} 
  & $\num{0.0926}$ {\scriptsize$\pm\num{0.0306}$} 
  & $\num{0.102}$ {\scriptsize$\pm\num{0.0143}$}
  \\
  \hline
\end{tabular}
\caption{Performance of the target models on the TUH-EEG and ELD datasets. Each entry reports the mean ± standard deviation across five runs with different random seeds for Train, Validation, and Test splits (unscaled metrics).
}
\label{tb:target_model}
\end{table*}

%% file: Results_includes/attacks_perfromance_eeg_table.tex
\begin{table*}[!htb]
\centering
\begin{tabular}{l l R{1.55cm} R{1.55cm} R{1.55cm} R{1.55cm} R{1.55cm} R{1.55cm}} 

\toprule

& \makecell[r]{} & \multicolumn{3}{c}{N-HiTS} & \multicolumn{3}{c}{LSTM} \\
\cmidrule(lr){3-5}
\cmidrule(lr){6-8}

& & \multicolumn{2}{c}{\footnotesize Record-level} & \footnotesize User-level & \multicolumn{2}{c}{\footnotesize Record-level} & \footnotesize User-level \\
\cmidrule(lr){3-4} \cmidrule(lr){5-5} \cmidrule(lr){6-7} \cmidrule(lr){8-8}
Attacks & Signal/Model & \footnotesize 0.1\%~FPR & \footnotesize 0.01\%~FPR & \footnotesize 0\%~FPR & \footnotesize 0.1\%~FPR & \footnotesize 0.01\%~FPR & \footnotesize 0\%~FPR \\
\midrule

\multirow{7}{*}{Ensemble}
& MSE          & 0.00 \spm{0.00}      & 0.00 \spm{0.00}      & \scl 6.00 \spm{7.35}          & 0.00 \spm{0.00}      & 0.00 \spm{0.00}      & \scl 2.00 \spm{2.45}          \\
& MAE          & 0.00 \spm{0.00}      & 0.00 \spm{0.00}      & \scl 8.00 \spm{8.12}          & 0.00 \spm{0.00}      & 0.00 \spm{0.00}      & \scl 7.00 \spm{5.10}          \\
& SMAPE        & 0.00 \spm{0.00}      & 0.00 \spm{0.00}      & \scl 6.00 \spm{2.00}          & 0.00 \spm{0.00}      & 0.00 \spm{0.00}      & \scl 5.00 \spm{5.48}          \\
& \rSMAPE      & 0.00 \spm{0.00}      & 0.00 \spm{0.00}      & \scl 8.00 \spm{4.00}          & 0.00 \spm{0.00}      & 0.00 \spm{0.00}      & \scl \B{10.00} \spm{7.07}     \\
& Trend        & \B{0.03} \spm{0.05}  & 0.00 \spm{0.00}      & \scl 4.00 \spm{5.83}          & \B{0.01} \spm{0.01}  & 0.00 \spm{0.00}      & \scl 5.00 \spm{4.47}          \\
& Seasonality  & \B{0.03} \spm{0.03}  & 0.00 \spm{0.00}      & \scl 9.00 \spm{8.00}          & 0.00 \spm{0.00}      & 0.00 \spm{0.00}      & \scl 4.00 \spm{4.90}          \\
& TS2Vec       & 0.00 \spm{0.00}      & 0.00 \spm{0.00}      & \scl \B{15.00} \spm{14.83}    & 0.00 \spm{0.00}      & 0.00 \spm{0.00}      & \scl 1.00 \spm{2.00}          \\

\midrule\midrule

\multirow{7}{*}{RMIA-Online}
& MSE          & 3.28 \spm{2.02}      & 1.38 \spm{1.98}      & \scl \colorbox{myblue}{\B{100.0} \spm{0.00}}     & 1.90 \spm{1.92}      & 0.16 \spm{0.24}      & \scl 32.00 \spm{26.94}        \\
& MAE          & \B{3.40} \spm{1.91}  & \B{1.48} \spm{1.93}  & \scl \colorbox{myblue}{\B{100.0} \spm{0.00}}     & \B{2.02} \spm{2.12}  & 0.36 \spm{0.37}      & \scl \colorbox{myblue}{\B{41.00} \spm{25.38}}    \\
& SMAPE        & 1.70 \spm{1.57}      & 0.46 \spm{0.53}      & \scl \colorbox{myblue}{\B{100.0} \spm{0.00}}     & 1.40 \spm{2.27}      & 1.06 \spm{2.11}      & \scl 36.00 \spm{23.54}        \\
& Trend        & 1.01 \spm{1.43}      & 0.38 \spm{0.68}      & \scl 58.00 \spm{21.12}        & 1.19 \spm{1.90}      & 0.58 \spm{1.11}      & \scl 10.00 \spm{7.07}         \\
& Seasonality  & 2.08 \spm{1.56}      & 0.74 \spm{1.09}      & \scl \colorbox{myblue}{\B{100.0} \spm{0.00}}     & 1.49 \spm{1.84}      & 0.56 \spm{0.90}      & \scl 30.00 \spm{21.68}        \\
& TS2Vec       & 2.61 \spm{1.80}      & 1.31 \spm{1.98}      & \scl \colorbox{myblue}{\B{100.0} \spm{0.00}}     & 1.84 \spm{2.30}      & \B{1.28} \spm{2.23}  & \scl 37.00 \spm{23.37}        \\

\midrule
 
\multirow{7}{*}{RMIA-Offline}
& MSE          & 1.24 \spm{2.25}      & 0.00 \spm{0.00}      & \scl \colorbox{myorange}{\B{42.00} \spm{30.59} }   & 1.51 \spm{1.24}      & 0.66 \spm{1.32}      & \scl 17.00 \spm{14.35}        \\
& MAE          & 1.10 \spm{2.06}      & \colorbox{myorange}{\B{1.04} \spm{2.09}}  & \scl 40.00 \spm{29.50}        & 1.67 \spm{1.17}      & 0.60 \spm{1.21}      & \scl 15.00 \spm{10.49}        \\
& SMAPE        & 2.20 \spm{1.58}      & 0.00 \spm{0.00}      & \scl 27.00 \spm{18.60}        & 1.24 \spm{1.39}      & 0.01 \spm{0.01}      & \scl 12.00 \spm{6.78}         \\
& Trend        & 0.46 \spm{0.28}      & 0.03 \spm{0.05}      & \scl 9.00 \spm{4.90}          & 0.38 \spm{0.41}      & 0.02 \spm{0.03}      & \scl 10.00 \spm{5.48}         \\
& Seasonality  & \colorbox{myorange}{\B{2.37} \spm{2.19}}  & 0.97 \spm{1.45}      & \scl 14.00 \spm{10.68}        & 1.87 \spm{1.90}      & \B{0.73} \spm{1.39}  & \scl 8.00 \spm{8.12}          \\
& TS2Vec       & 0.29 \spm{0.59}      & 0.00 \spm{0.00}      & \scl 41.00 \spm{25.38}        & \B{1.94} \spm{1.27}  & 0.00 \spm{0.00}      & \scl \colorbox{myorange}{\B{19.00} \spm{12.41}}   \\

\midrule\midrule

\multirow{8}{*}{LiRA-Online}
& MSE          & 4.02 \spm{2.81}      & 0.19 \spm{0.24}      & \scl \colorbox{myblue}{\B{100.0} \spm{0.00}}     & 3.05 \spm{2.35}      & 2.27 \spm{2.03}      & \scl 22.00 \spm{10.77}            \\
& MAE          & 6.73 \spm{2.60}      & 1.35 \spm{1.79}      & \scl \colorbox{myblue}{\B{100.0} \spm{0.00}}    & 3.43 \spm{2.47}      & 2.58 \spm{2.19}      & \scl 28.00 \spm{15.03}            \\
& SMAPE        & 5.04 \spm{2.20}      & 1.11 \spm{1.10}      & \scl \colorbox{myblue}{\B{100.0} \spm{0.00}}   & 3.19 \spm{2.37}      & 1.93 \spm{2.08}      & \scl 34.00 \spm{12.41}            \\
& \rSMAPE      & 6.65 \spm{2.39}      & 2.26 \spm{1.83}      & \scl \colorbox{myblue}{\B{100.0} \spm{0.00}}   & 3.50 \spm{2.40}      & 2.23 \spm{2.13}      & \scl \B{35.00} \spm{13.42}        \\
& Trend        & 5.19 \spm{4.12}      & 2.42 \spm{2.05}      & \scl 94.00 \spm{4.90}         & 2.26 \spm{2.51}      & 1.99 \spm{2.39}      & \scl 15.00 \spm{5.48}             \\
& Seasonality  & 7.28 \spm{2.50}      & 1.77 \spm{2.12}      & \scl \colorbox{myblue}{\B{100.0} \spm{0.00}}     & 3.42 \spm{2.41}      & \B{2.61} \spm{2.13}  & \scl 23.00 \spm{12.08}            \\
& TS2Vec       & \B{7.41} \spm{0.97}  & \B{2.64} \spm{1.49}  & \scl \colorbox{myblue}{\B{100.0} \spm{0.00}}    & \B{3.57} \spm{2.19}  & 2.57 \spm{2.19}      & \scl 31.00 \spm{9.70}             \\

\midrule

\multirow{7}{*}{LiRA-Offline}
& MSE            & 0.01 \spm{0.02}      & 0.00 \spm{0.00}      & \scl 2.00 \spm{2.45}          & 1.66 \spm{2.16}      & 1.35 \spm{2.13}      & \scl 4.00 \spm{4.90}              \\
& MAE            & 0.07 \spm{0.10}      & 0.00 \spm{0.00}      & \scl 6.00 \spm{8.00}          & 1.97 \spm{2.35}      & \B{1.80} \spm{2.29}  & \scl 6.00 \spm{3.74}              \\
& SMAPE          & 0.20 \spm{0.11}      & 0.05 \spm{0.03}      & \scl 16.00 \spm{5.83}         & 1.88 \spm{2.25}      & 1.54 \spm{1.90}      & \scl \B{8.00} \spm{6.00}          \\
& \rSMAPE        & \B{1.04} \spm{0.93}  & 0.12 \spm{0.08}      & \scl \B{22.00} \spm{9.80}     & \colorbox{myorange}{\B{2.12} \spm{2.38}}  & 1.69 \spm{1.98}      & \scl \B{8.00} \spm{4.00}          \\
& Trend          & 0.55 \spm{0.57}      & \B{0.13} \spm{0.14}  & \scl 12.00 \spm{8.72}         & 1.84 \spm{2.14}      & 1.46 \spm{1.98}      & \scl 3.00 \spm{4.00}              \\
& Seasonality    & 0.08 \spm{0.11}      & 0.01 \spm{0.01}      & \scl 10.00 \spm{7.07}         & 1.97 \spm{2.34}      & 1.71 \spm{2.22}      & \scl 6.00 \spm{3.74}              \\
& TS2Vec         & 0.33 \spm{0.34}      & 0.05 \spm{0.03}      & \scl 8.00 \spm{6.78}          & 1.90 \spm{2.24}      & 1.71 \spm{2.22}      & \scl 5.00 \spm{3.16}              \\

\midrule\midrule

{Multi-Signal LiRA-Online}
&   & 9.11 \spm{2.86}      & 2.71 \spm{2.88}      & \scl \colorbox{myblue}{{100.0} \spm{0.00} }       & \colorbox{myblue}{{3.79} \spm{2.51} }     & \colorbox{myblue}{2.99 \spm{2.35}}      & \scl 32.00 \spm{8.72}          \\

\midrule

{Multi-Signal LiRA-Offline}
&   & 0.12 \spm{0.16}      & 0.01 \spm{0.00}      & \scl 10.00 \spm{8.94}         & \colorbox{myorange}{{2.12} \spm{2.46}}     & \colorbox{myorange}{{1.88} \spm{2.35}}      & \scl 5.00 \spm{3.16}        \\

\midrule\midrule
\multirow{2}{*}{{DTS-Online}}
& LSTM            & \colorbox{myblue}{\B{23.32} \spm{6.65}}  & 15.62 \spm{7.22}     & \scl \B{96.00} \spm{4.90}     & \B{2.97} \spm{3.75}  & \B{2.15} \spm{2.69}  & \scl \B{30.00} \spm{23.45}    \\ 
& InceptionTime  & 22.39 \spm{7.80}     & \colorbox{myblue}{\B{15.81} \spm{7.35}} & \scl 90.00 \spm{6.32}         & 2.49 \spm{4.99}      & 1.70 \spm{3.39}      & \scl 20.00 \spm{19.24}        \\

\midrule

\multirow{2}{*}{{DTS-Offline}}
& LSTM          & \B{2.09} \spm{3.13}  & \B{0.67} \spm{1.01}  & \scl \B{27.00} \spm{16.61}    & 0.17 \spm{0.16}      & 0.00 \spm{0.01}      & \scl \B{9.00} \spm{15.62}     \\     
& InceptionTime & 1.12 \spm{1.45}      & 0.64 \spm{1.28}      & \scl 22.00 \spm{23.79}        & \B{0.23} \spm{0.29}  & \B{0.05} \spm{0.10}  & \scl 5.00 \spm{7.75}          \\

\bottomrule
\end{tabular}
\caption{Membership inference attack performance on the TUH-EEG dataset. We report record-level TPR at fixed FPRs (0.1\% and 0.01\%) and user-level TPR at 0\% FPR. Results are averaged over five runs, with mean and standard deviation shown. For attacks with multiple configurations, the best-performing configuration is bolded. 
The \colorbox{myblue}{\phantom{XX}} and \colorbox{myorange}{\phantom{XX}} indicate the best-performing attack over each metric in the online and offline setting, respectively.}
\label{tab:eeg_results_summary}
\end{table*}

%% file: Results_includes/attacks_perfromance_eld_table.tex
\begin{table*}[!htb]
\centering
\begin{tabular}{l l R{1.55cm} R{1.55cm} R{1.55cm} R{1.55cm} R{1.55cm} R{1.55cm}} 

\toprule

& \makecell[r]{} & \multicolumn{3}{c}{N-HiTS} & \multicolumn{3}{c}{LSTM} \\
\cmidrule(lr){3-5}
\cmidrule(lr){6-8}

& & \multicolumn{2}{c}{\footnotesize Record-level} & \footnotesize User-level & \multicolumn{2}{c}{\footnotesize Record-level} & \footnotesize User-level \\
\cmidrule(lr){3-4} \cmidrule(lr){5-5} \cmidrule(lr){6-7} \cmidrule(lr){8-8}
Attacks & Signal/Model & \footnotesize 0.1\%~FPR & \footnotesize 0.01\%~FPR & \footnotesize 0\%~FPR & \footnotesize 0.1\%~FPR & \footnotesize 0.01\%~FPR & \footnotesize 0\%~FPR \\
\midrule


\multirow{7}{*}{Ensemble Attack}
& MSE             & 0.01 \spm{0.02}      & 0.00 \spm{0.00}      & \scl 8.00 \spm{6.78}          & 0.00 \spm{0.00}      & 0.00 \spm{0.00}      & \scl 4.00 \spm{4.90}          \\
& MAE          & 0.00 \spm{0.00}      & 0.00 \spm{0.00}      & \scl 6.00 \spm{7.35}          & 0.00 \spm{0.00}      & 0.00 \spm{0.00}      & \scl 5.00 \spm{5.48}          \\
& SMAPE        & \B{0.02} \spm{0.02}  & 0.00 \spm{0.00}      & \scl 2.00 \spm{4.00}          & \B{0.02} \spm{0.05}  & 0.00 \spm{0.00}      & \scl 3.00 \spm{6.00}          \\
& \rSMAPE      & 0.00 \spm{0.00}      & 0.00 \spm{0.00}      & \scl 3.00 \spm{4.00}          & 0.00 \spm{0.00}      & 0.00 \spm{0.00}      & \scl 0.00 \spm{0.00}          \\
& Trend        & 0.01 \spm{0.02}      & 0.00 \spm{0.00}      & \scl 5.00 \spm{4.47}          & 0.00 \spm{0.00}      & 0.00 \spm{0.00}      & \scl 2.00 \spm{2.45}          \\
& Seasonality  & 0.00 \spm{0.00}      & 0.00 \spm{0.00}      & \scl \B{9.00} \spm{9.70}      & 0.00 \spm{0.00}      & 0.00 \spm{0.00}      & \scl \B{6.00} \spm{3.74}      \\
& TS2Vec       & 0.00 \spm{0.00}      & 0.00 \spm{0.00}      & \scl 2.00 \spm{4.00}          & 0.00 \spm{0.00}      & 0.00 \spm{0.00}      & \scl 3.00 \spm{4.00}          \\

\midrule\midrule

\multirow{6}{*}{RMIA-Online}
& MSE           & 1.12 \spm{1.26}      & \B{0.31} \spm{0.37}  & \scl 44.00 \spm{20.35}        & \B{1.16} \spm{0.58}  & \B{0.15} \spm{0.14}  & \scl \B{19.00} \spm{17.44}      \\
& MAE           & \B{1.19} \spm{1.20}  & 0.26 \spm{0.40}      & \scl 59.00 \spm{19.60}        & 1.00 \spm{0.60}      & 0.13 \spm{0.15}      & \scl 16.00 \spm{12.41}          \\
& SMAPE         & 0.67 \spm{0.44}      & 0.09 \spm{0.12}      & \scl 47.00 \spm{21.59}        & 0.24 \spm{0.21}      & 0.04 \spm{0.05}      & \scl 16.00 \spm{11.58}          \\
& Trend         & 0.17 \spm{0.12}      & 0.02 \spm{0.04}      & \scl 21.00 \spm{7.35}         & 0.09 \spm{0.06}      & 0.01 \spm{0.01}      & \scl 13.00 \spm{11.66}          \\
& Seasonality   & 0.97 \spm{0.69}      & 0.04 \spm{0.05}      & \scl \colorbox{myblue}{\B{73.00} \spm{19.65}}    & 0.46 \spm{0.37}      & 0.08 \spm{0.10}      & \scl \B{19.00} \spm{14.97}      \\
& TS2Vec        & 0.79 \spm{0.50}      & 0.03 \spm{0.05}      & \scl 64.00 \spm{30.23}        & 0.40 \spm{0.47}      & 0.10 \spm{0.19}      & \scl 18.00 \spm{11.66}          \\

\midrule
 
\multirow{6}{*}{RMIA-Offline}
& MSE          & 0.06 \spm{0.13}      & 0.00 \spm{0.00}      & \scl 7.00 \spm{6.00}          & 1.28 \spm{1.33}      & 0.26 \spm{0.23}      & \scl 6.00 \spm{5.83}          \\
& MAE          & 0.50 \spm{0.66}      & 0.00 \spm{0.01}      & \scl \B{8.00} \spm{6.78}      & \colorbox{myorange}{\B{1.29} \spm{1.44}}  & 0.03 \spm{0.06}      & \scl 5.00 \spm{6.32}          \\
& SMAPE        & 0.02 \spm{0.04}      & 0.00 \spm{0.00}      & \scl \B{8.00} \spm{2.45}      & 0.61 \spm{0.45}      & 0.00 \spm{0.00}      & \scl 4.00 \spm{4.90}          \\
& Trend        & 0.34 \spm{0.15}      & 0.03 \spm{0.06}      & \scl 6.00 \spm{5.83}          & 0.16 \spm{0.09}      & 0.02 \spm{0.03}      & \scl 6.00 \spm{5.83}          \\
& Seasonality  & \colorbox{myorange}{\B{0.89} \spm{0.86}}  & \colorbox{myorange}{\B{0.31} \spm{0.50}}  & \scl 7.00 \spm{6.00}          & 0.68 \spm{1.02}      & \colorbox{myorange}{\B{0.42} \spm{0.84}}  & \scl 5.00 \spm{6.32}          \\
& TS2Vec       & 0.61 \spm{0.90}      & 0.03 \spm{0.04}      & \scl 7.00 \spm{5.10}          & 0.77 \spm{1.03}      & 0.36 \spm{0.73}      & \scl \B{7.00} \spm{7.48}      \\

\midrule\midrule

\multirow{7}{*}{LiRA-Online}
& MSE           & 1.06 \spm{0.57}      & 0.28 \spm{0.21}      & \scl 43.00 \spm{28.21}        & 2.26 \spm{3.18}      & 0.84 \spm{1.19}      & \scl 29.00 \spm{20.10}         \\
& MAE           & 1.99 \spm{1.00}      & 0.65 \spm{0.36}      & \scl 49.00 \spm{26.72}        & 3.29 \spm{2.64}      & 1.47 \spm{0.88}      & \scl 35.00 \spm{15.49}         \\
& SMAPE         & 1.50 \spm{0.61}      & 0.46 \spm{0.23}      & \scl 43.00 \spm{26.76}        & 2.18 \spm{1.46}      & 0.78 \spm{0.29}      & \scl 29.00 \spm{11.58}         \\
& \rSMAPE       & 1.94 \spm{0.79}      & 0.59 \spm{0.26}      & \scl \B{53.00} \spm{29.43}    & 3.10 \spm{1.75}      & 0.93 \spm{0.51}      & \scl \colorbox{myblue}{\B{38.00} \spm{12.08}}     \\
& Trend         & 1.34 \spm{0.29}      & 0.45 \spm{0.20}      & \scl 50.00 \spm{19.49}        & 1.17 \spm{0.72}      & 0.31 \spm{0.32}      & \scl 27.00 \spm{19.65}         \\
& Seasonality   & 2.02 \spm{0.85}      & 0.76 \spm{0.37}      & \scl 49.00 \spm{26.72}        & 3.46 \spm{3.07}      & 1.43 \spm{1.14}      & \scl 34.00 \spm{13.56}         \\
& TS2Vec        & \B{2.16} \spm{1.08}  & \B{0.88} \spm{0.45}  & \scl 48.00 \spm{25.42}        & \B{3.50} \spm{2.86}  & \B{1.68} \spm{1.25}  & \scl 33.00 \spm{12.08}         \\

\midrule

\multirow{7}{*}{LiRA-Offline}
& MSE            & 0.07 \spm{0.07}      & 0.01 \spm{0.01}      & \scl 4.00 \spm{5.83}          & 0.03 \spm{0.03}      & 0.00 \spm{0.00}      & \scl 3.00 \spm{4.00}             \\
& MAE            & 0.26 \spm{0.17}      & 0.07 \spm{0.07}      & \scl 9.00 \spm{7.35}          & 0.21 \spm{0.24}      & 0.02 \spm{0.02}      & \scl 7.00 \spm{4.00}             \\
& SMAPE          & 0.35 \spm{0.23}      & 0.09 \spm{0.06}      & \scl 10.00 \spm{7.07}         & 0.18 \spm{0.14}      & 0.05 \spm{0.06}      & \scl 8.00 \spm{7.48}             \\
& \rSMAPE        & \B{0.59} \spm{0.27}  & 0.13 \spm{0.10}      & \scl \B{12.00} \spm{9.27}     & 0.34 \spm{0.26}      & 0.07 \spm{0.04}      & \scl \colorbox{myorange}{\B{9.00} \spm{4.90}}         \\
& Trend          & 0.22 \spm{0.10}      & 0.05 \spm{0.03}      & \scl 6.00 \spm{3.74}          & 0.06 \spm{0.08}      & 0.01 \spm{0.02}      & \scl 5.00 \spm{4.47}             \\
& Seasonality    & 0.30 \spm{0.18}      & 0.08 \spm{0.09}      & \scl 10.00 \spm{8.37}         & 0.24 \spm{0.18}      & 0.07 \spm{0.09}      & \scl 8.00 \spm{5.10}             \\
& TS2Vec         & 0.52 \spm{0.36}      & \B{0.26} \spm{0.23}  & \scl 9.00 \spm{5.83}          & \B{0.41} \spm{0.32}  & \B{0.10} \spm{0.06}  & \scl 7.00 \spm{5.10}             \\

\midrule\midrule


Multi-Signal LiRA-Online
& & 2.32 \spm{1.01}      & 0.85 \spm{0.41}      & \scl 48.00 \spm{26.76}        & 4.15 \spm{4.13}      & 2.29 \spm{2.27}      & \scl 34.00 \spm{12.81}           \\
\midrule

Multi-Signal LiRA-Offline
& & 0.32 \spm{0.20}      & 0.08 \spm{0.08}      & \scl 10.00 \spm{6.32}         & 0.37 \spm{0.42}      & 0.06 \spm{0.06}      & \scl 8.00 \spm{4.00}              \\

\midrule\midrule
\multirow{2}{*}{DTS-Online}
& LSTM           & 6.34 \spm{3.91}      & 2.77 \spm{1.87}      & \scl 56.00 \spm{22.45}        & \colorbox{myblue}{\B{4.78} \spm{4.56}}  & \colorbox{myblue}{\B{2.57} \spm{2.42}}  & \scl \B{34.00} \spm{21.54}    \\
& InceptionTime  & \colorbox{myblue}{\B{6.66} \spm{3.48}}  & \colorbox{myblue}{\B{2.94} \spm{1.69}}  & \scl \B{65.00} \spm{21.21}    & 4.49 \spm{4.83}      & 2.40 \spm{3.20}      & \scl 24.00 \spm{25.77}        \\

\midrule

\multirow{2}{*}{DTS-Offline}
& LSTM            & 0.57 \spm{0.39}      & 0.15 \spm{0.18}      & \scl \colorbox{myorange}{\B{13.00} \spm{12.88}}    & 0.76 \spm{0.85}      & 0.26 \spm{0.39}      & \scl \colorbox{myorange}{\B{9.00} \spm{5.83}}      \\
& InceptionTime   & \B{0.68} \spm{0.71}  & \B{0.21} \spm{0.26}  & \scl 11.00 \spm{13.19}        & \B{0.78} \spm{0.86}  & \B{0.30} \spm{0.47}  & \scl 6.00 \spm{4.90}          \\

\bottomrule
\end{tabular}
\caption{Membership inference attack performance on the ELD dataset. We report record-level TPR at fixed FPRs (0.1\% and 0.01\%) and user-level TPR at 0\% FPR. Results are averaged over five runs, with mean and standard deviation shown. For attacks with multiple configurations, the best-performing configuration is bolded. 
The \colorbox{myblue}{\phantom{XX}} and \colorbox{myorange}{\phantom{XX}} indicate the best-performing attack over each metric in the online and offline setting, respectively.}

\label{tab:eld_results_summary}
\end{table*}

%% file: Results_includes/ROCs_AUC_table.tex
\begin{table*}[!htb]

\centering
\begin{tabular}{l l R{2cm} R{2cm} R{2cm} R{2cm}}
\toprule
& \makecell[r]{} 
& \multicolumn{2}{c}{TUH-EEG} 
& \multicolumn{2}{c}{ELD} \\
\cmidrule(lr){3-4} \cmidrule(lr){5-6}
Attacks & Signal/Model 
& \footnotesize N-HiTS & \footnotesize LSTM 
& \footnotesize N-HiTS & \footnotesize LSTM \\
\midrule
\multirow{7}{*}{Ensemble Attack}
& MSE          & 0.546 \spm{0.0823}       & 0.530 \spm{0.0420}    &  0.504 \spm{0.0908}       & \B{0.509} \spm{0.0764}   \\
& MAE          & 0.532 \spm{0.0909}       & \B{0.532} \spm{0.0525} & 0.503 \spm{0.0709}       & 0.500 \spm{0.0767} \\
& SMAPE        & 0.491 \spm{0.0476}       & 0.495 \spm{0.0312}    & 0.496 \spm{0.0377}       & 0.482 \spm{0.0711}  \\
& \rSMAPE      & 0.494 \spm{0.0454}       & 0.499 \spm{0.0392}    & 0.493 \spm{0.0385}       & 0.484 \spm{0.0648}  \\
& Trend        & 0.537 \spm{0.0291}       & 0.506 \spm{0.0300}    & \B{0.509} \spm{0.0622}   & 0.496 \spm{0.0599}    \\
& Seasonality  & 0.538 \spm{0.0911}       & 0.529 \spm{0.0483}     & 0.484 \spm{0.0905}       & 0.502 \spm{0.0719}  \\
& TS2Vec       & \B{0.555} \spm{0.1110}   & 0.507 \spm{0.0795}     &  0.477 \spm{0.0599}       & 0.499 \spm{0.0750} \\

\midrule\midrule
\multirow{6}{*}{RMIA-Online}
& MSE          & 0.744 \spm{0.0201}       & 0.602 \spm{0.0302}       & 0.646 \spm{0.0454}       & 0.658 \spm{0.0362}       \\
& MAE          & 0.775 \spm{0.0142}       & 0.614 \spm{0.0384}       & \B{0.665} \spm{0.0470}   & \B{0.669} \spm{0.0322}   \\
& SMAPE        & 0.747 \spm{0.0089}      & \B{0.614} \spm{0.0233}   & 0.628 \spm{0.0554}       & 0.636 \spm{0.0207}       \\
& Trend        & 0.610 \spm{0.0121}       & 0.527 \spm{0.0122}       & 0.581 \spm{0.0190}       & 0.559 \spm{0.00678}      \\
& Seasonality  & 0.778 \spm{0.0282}       & 0.594 \spm{0.0593}       & 0.660 \spm{0.0483}       & 0.666 \spm{0.0312}       \\
& TS2Vec       & \B{0.782} \spm{0.0173}   & 0.609 \spm{0.0515}       & 0.648 \spm{0.0414}       & 0.654 \spm{0.0316}       \\

\midrule
\multirow{6}{*}{RMIA-Offline}
& MSE          & 0.724 \spm{0.0392}       & 0.631 \spm{0.0395}       & 0.592 \spm{0.0326}       & 0.593 \spm{0.0466}       \\
& MAE          & 0.744 \spm{0.0443}       & \colorbox{myorange}{\B{0.637} \spm{0.0406}}   & 0.593 \spm{0.0515}       & 0.580 \spm{0.0498}       \\
& SMAPE        & 0.665 \spm{0.0334}       & 0.588 \spm{0.0265}       & 0.600 \spm{0.0135}       & 0.593 \spm{0.0358}       \\
& Trend        & 0.588 \spm{0.0203}       & 0.551 \spm{0.0267}       & 0.550 \spm{0.0534}       & 0.535 \spm{0.0394}       \\
& Seasonality  & 0.706 \spm{0.0423}       & 0.622 \spm{0.0262}       & 0.558 \spm{0.0767}       & 0.537 \spm{0.0699}       \\
& TS2Vec       & \colorbox{myorange}{\B{0.748} \spm{0.0488}}   & 0.634 \spm{0.0416}       & \colorbox{myorange}{\B{0.607} \spm{0.0223}}   & \colorbox{myorange}{\B{0.599} \spm{0.0360}}   \\

\midrule\midrule

\multirow{7}{*}{LiRA-Online}
& MSE          & 0.811 \spm{0.0099}       & 0.640 \spm{0.0333}       & 0.672 \spm{0.0445}       & 0.672 \spm{0.0519}       \\
& MAE          & 0.816 \spm{0.0102}        & 0.646 \spm{0.0306}       & 0.682 \spm{0.0433}       & \B{0.681} \spm{0.0444}   \\
& SMAPE        & 0.796 \spm{0.0089}       & 0.647 \spm{0.0186}       & 0.660 \spm{0.0366}       & 0.661 \spm{0.0366}       \\
& \rSMAPE      & 0.801 \spm{0.0100}       & 0.650 \spm{0.0202}       & 0.668 \spm{0.0361}       & 0.666 \spm{0.0297}       \\
& Trend        & 0.697 \spm{0.0204}        & 0.552 \spm{0.0159}       & 0.646 \spm{0.0169}       & 0.602 \spm{0.0397}       \\
& Seasonality  & 0.821 \spm{0.0112}        & 0.646 \spm{0.0315}       & \B{0.682} \spm{0.0437}   & 0.680 \spm{0.0437}       \\
& TS2Vec       & \B{0.822} \spm{0.0147}    & \B{0.655} \spm{0.0374}   & 0.669 \spm{0.0434}       & 0.672 \spm{0.0495}       \\

\midrule

\multirow{7}{*}{LiRA-Offline}
& MSE           & 0.440 \spm{0.0284}       & 0.445 \spm{0.0321}       & 0.510 \spm{0.0689}       & 0.461 \spm{0.0840}       \\
& MAE           & 0.495 \spm{0.0288}       & 0.474 \spm{0.0355}       & 0.523 \spm{0.0558}       & 0.469 \spm{0.0780}       \\
& SMAPE         & 0.521 \spm{0.0196}       & 0.486 \spm{0.0257}       & 0.495 \spm{0.0146}       & 0.461 \spm{0.0398}       \\
& \rSMAPE       & \B{0.563} \spm{0.0213}   & \B{0.518} \spm{0.0304}   & \B{0.532} \spm{0.0149}   & \B{0.488} \spm{0.0414}   \\
& Trend         & 0.470 \spm{0.0364}       & 0.459 \spm{0.0548}       & 0.492 \spm{0.0654}       & 0.477 \spm{0.0632}       \\
& Seasonality   & 0.495 \spm{0.0305}       & 0.476 \spm{0.0381}       & 0.521 \spm{0.0568}       & 0.469 \spm{0.0833}       \\
& TS2Vec        & 0.506 \spm{0.0182}       & 0.480 \spm{0.0231}       & 0.500 \spm{0.0331}       & 0.464 \spm{0.0405}       \\

\midrule\midrule

\multirow{1}{*}{Multi-Signal LiRA-Online}
&            &  0.847 \spm{0.0119}        &  \colorbox{myblue}{0.661 \spm{0.0312}}  & 0.703 \spm{0.0439}       &   \colorbox{myblue}{ 0.700 \spm{0.0464}}    \\

\midrule

\multirow{1}{*}{Multi-Signal LiRA-Offline}
&            &  0.492 \spm{0.0278}         & 0.469 \spm{0.0378} & 0.517 \spm{0.0567}       &    0.459 \spm{0.0812}   \\

\midrule\midrule
\multirow{2}{*}{DTS-Online}
& LSTM           & \colorbox{myblue}{\B{0.884} \spm{0.0303}}   & \B{0.637} \spm{0.0378} & 0.754 \spm{0.0131}& \B{0.667} \spm{0.069}  \\
& InceptionTime  & 0.866 \spm{0.0389}       & 0.597 \spm{0.0540} & \colorbox{myblue}{\B{0.758} \spm{0.0137}}& 0.658 \spm{0.081}      \\

\midrule

\multirow{2}{*}{DTS-Offline}
& LSTM           & 0.653 \spm{0.0288}       & \B{0.529} \spm{0.0355}  &  \B{0.574} \spm{0.0395} &  0.549 \spm{0.0312}\\
& InceptionTime  & \B{0.660} \spm{0.0503}   & 0.504 \spm{0.0507}  &  0.56 \spm{0.0218}  &\B{0.55} \spm{0.0265} \\

\bottomrule
\end{tabular}

\caption{Membership inference attack performance on the TUH-EEG and ELD dataset. We report the ROC-AUC averaged over five runs, with mean and standard deviation shown. For attacks with multiple configurations, the best-performing configuration is bolded. 
The \colorbox{myblue}{\phantom{XX}} and \colorbox{myorange}{\phantom{XX}} indicate the best-performing attack in the online and offline setting, respectively.}

\label{tab:rocauc}
\end{table*}

%% file: Results_includes/fig_horizon.tex
\pgfplotstableread{
Horizon   sample_fpr_01    sample_fpr_001     user_fpr_0
5         1.4826  0.3310  80.0
10        2.8081  0.6128  35.0
20        3.7155  1.5379  60.0
40        2.1005  0.8307  15.0
80        3.2717  1.3245  15.0
}\lira

\pgfplotstableread{
Horizon   sample_fpr_01         sample_fpr_001         user_fpr_0
5         1.4480397422 0.1792427497 15.0
10        2.7734873413 0.7900745417 35.0
20        3.4654929104 1.4746992810 45.0
40        2.3464100666 0.7001547675 15.0
80        4.0375143378 1.6709398826 15.0
}\mslira

\pgfplotstableread{
Horizon   sample_fpr_01         sample_fpr_001         user_fpr_0
5         0.5095327605 0.0057062299 50.0
10        2.2322208045 0.4183735142 60.0
20        3.3018614340 1.6272427928 85.0
40        8.8819729493 4.1615638248 80.0
80        9.0371769786 5.1511369003 85.0
}\dtsmia
\begin{tikzpicture}
    \begin{groupplot}[
        group style={
            group size=3 by 1, 
            horizontal sep=2cm,
        },
        width=4.5cm,
        height=4.5cm,
        grid=both, 
        legend style={
            legend columns=3, 
            /tikz/every even column/.append style={column sep=1em}, 
        },
    ]
\nextgroupplot[
    xmax=80,
    xmin=5,
    ylabel={$\text{TPR} @ \text{FPR} = 0.1\%$},
    xlabel={Horizon},
    ymin=0,
    ymax=10,
    xtick={5, 10, 20, 40, 80},
    xticklabels = {$5$, $10$, $20$, $40$, $80$},
    grid=major,
    legend to name=sharedlegend,
    xmode=log
]

    \addplot+ table[x=Horizon, y=sample_fpr_01] {\lira};
    \addlegendentry{LiRA-Online (TS2Vec)}

    \addplot+ table[x=Horizon, y=sample_fpr_01] {\mslira};
    \addlegendentry{Multi-Signal LiRA-Online}

    \addplot[color=brown, mark=*] table[x=Horizon, y=sample_fpr_01] {\dtsmia};
    \addlegendentry{DTS (InceptionTime)}
\vspace{-2cm}
\nextgroupplot[
        xmax=80,
        xmin=5,
        ymin=0,
        ymax=6,
        ylabel={$\text{TPR} @ \text{FPR} = 0.01\%$},
        xlabel={Horizon},
        xtick={5, 10, 20, 40, 80},
        xticklabels = {$5$, $10$, $20$, $40$, $80$},
        grid=major,
        xmode=log
        ]

    \addplot+ table[x=Horizon, y=sample_fpr_001] {\lira};
    \addplot+ table[x=Horizon, y=sample_fpr_001] {\mslira};
    \addplot[color=brown, mark=*] table[x=Horizon, y=sample_fpr_001] {\dtsmia};
    
    \nextgroupplot[
        xmax=80,
        xmin=5,
        ymin=0,
        ymax=100,
        ylabel={$\text{TPR}_{\text{user}} @ \text{FPR} = 0\%$},
        xtick={5, 10, 20, 40, 80},
        xticklabels = {$5$, $10$, $20$, $40$, $80$},
        xlabel={Horizon},
        xmode=log
    ]

    \addplot+ table[x=Horizon, y=user_fpr_0] {\lira};
    \addplot+ table[x=Horizon, y=user_fpr_0] {\mslira};
    \addplot[color=brown, mark=*] table[x=Horizon, y=user_fpr_0] {\dtsmia};

    \end{groupplot}
  \node at ($(group c1r1.north east)!0.5!(group c3r1.north west)+ (0, 0.5cm)$) {\pgfplotslegendfromname{sharedlegend}};
\end{tikzpicture}

%% file: Results_includes/fig_individuals.tex
\pgfplotstableread{
num   mean    std    
50         2.932464216114509  1.152793412252717 
100        2.155500302399032  1.082368109661184  
200        1.964686513003158  0.5455046240310213
}\liraA

\pgfplotstableread{
num   mean    std    
50         1.0953564948592165  1.1294707485197824
100        0.8771587930918621  0.45471520917495756
200        0.6859082050937437  0.34094460738778043
}\liraB

\pgfplotstableread{
num   mean    std    
50         62 20.39607805437114
100        48  25.41653005427767 
200        43  2.915475947422648
}\liraC

\pgfplotstableread{
num   mean    std    
50         4.21288891875546  1.5909669310610313
100       2.3220885693165783  1.0055861684548844
200       2.539211074524561  0.6343689206905576
}\msliraA

\pgfplotstableread{
num   mean    std    
50        1.4359250050399837  1.2409962721449592
100        0.8543108662052281  0.41061617944300954
200        0.7465560110207647  0.4107141430297063
}\msliraB

\pgfplotstableread{
num   mean    std    
50         70 17.888543819998322
100        48  26.758176320519304
200        41.5  7.516648189186453
}\msliraC

\pgfplotstableread{
num   mean    std    
50         8.238962435320207 3.334247325302559	
100       6.65533230293663 3.475666008129155
200       	3.499227202472952  2.008177480536386
}\dtsA

\pgfplotstableread{
num   mean    std    
50        3.1964249714400914  1.9851719399934515 
100        2.9371682010617564  1.6878727800447901
200        1.4782272696727372  0.9967583181245594
}\dtsB

\pgfplotstableread{
num   mean    std    
50         92 11.661903789690603
100        65  21.213203435596427
200        21  12.0
}\dtsC

\begin{tikzpicture}
    \begin{groupplot}[
        group style={
            group size=3 by 1, 
            horizontal sep=2cm,
        },
        width=4.5cm,
        height=4.5cm,
        grid=both, 
        legend style={
            legend columns=3, 
            /tikz/every even column/.append style={column sep=1em}, 
        },
    ]
\nextgroupplot[
    xmax=200,
    xmin=50,
    ylabel={$\text{TPR} @ \text{FPR} = 0.1\%$},
    xlabel={Number of individuals},
    ymin=0,
    ymax=12,
    grid=major,
    legend to name=sharedlegend,
]

    \addplot[name path=U, draw=none, mark=none, forget plot]table[x=num, y expr=\thisrow{mean} + \thisrow{std}] {\liraA};
    \addplot[name path=L, draw=none, mark = none, forget plot] table[x=num, y expr=\thisrow{mean} - \thisrow{std}] {\liraA};
    \addplot+[fill=blue, fill opacity=0.20, draw=none, forget plot] fill between[of=U and L];
    \addplot+[color=blue, solid, mark=*] table[x=num, y=mean]{\liraA};
    
    \addlegendentry{LiRA-Online (TS2Vec)}

    \addplot[name path=U, draw=none, mark=none, forget plot]table[x=num, y expr=\thisrow{mean} + \thisrow{std}] {\msliraA};
    \addplot[name path=L, draw=none, mark = none, forget plot] table[x=num, y expr=\thisrow{mean} - \thisrow{std}] {\msliraA};
    \addplot[fill=red, fill opacity=0.20, draw=none, forget plot] fill between[of=U and L];
    \addplot+[color=red,solid, mark=*] table[x=num, y=mean]{\msliraA};
    
    \addlegendentry{Multi-Signal LiRA-Online}

    \addplot[name path=U, draw=none, mark=none, forget plot]table[x=num, y expr=\thisrow{mean} + \thisrow{std}] {\dtsA};
    \addplot[name path=L, draw=none, mark = none, forget plot] table[x=num, y expr=\thisrow{mean} - \thisrow{std}] {\dtsA};
    \addplot+[fill=brown, fill opacity=0.20, draw=none, forget plot] fill between[of=U and L];
    \addplot+[color=brown, solid, mark=*] table[x=num, y=mean]{\dtsA};
    
    \addlegendentry{DTS (InceptionTime)}

\nextgroupplot[
        xmax=200,
        xmin=50,
        ymin=0,
        ymax=6,
        ylabel={$\text{TPR} @ \text{FPR} = 0.01\%$},
        xlabel={Number of individuals},
        grid=major,
        ]

    \addplot[name path=U, draw=none, mark=none]table[x=num, y expr=\thisrow{mean} + \thisrow{std}] {\liraB};
    \addplot[name path=L, draw=none, mark = none] table[x=num, y expr=\thisrow{mean} - \thisrow{std}] {\liraB};
    \addplot+[fill=blue, fill opacity=0.20, draw=none, forget plot] fill between[of=U and L];
    \addplot+[color=blue, solid, mark=*] table[x=num, y=mean]{\liraB};

    \addplot[name path=U, draw=none, mark=none]table[x=num, y expr=\thisrow{mean} + \thisrow{std}] {\msliraB};
    \addplot[name path=L, draw=none, mark = none] table[x=num, y expr=\thisrow{mean} - \thisrow{std}] {\msliraB};
    \addplot[fill=red, fill opacity=0.20, draw=none, forget plot] fill between[of=U and L];
    \addplot+[color=red, solid, mark=*] table[x=num, y=mean]{\msliraB};    

    \addplot[name path=U, draw=none, mark=none]table[x=num, y expr=\thisrow{mean} + \thisrow{std}] {\dtsB};
    \addplot[name path=L, draw=none, mark = none] table[x=num, y expr=\thisrow{mean} - \thisrow{std}] {\dtsB};
    \addplot+[fill=brown, fill opacity=0.20, draw=none, forget plot] fill between[of=U and L];
    \addplot+[color=brown, solid, mark=*] table[x=num, y=mean]{\dtsB};
    
    \nextgroupplot[
        xmax=200,
        xmin=50,
        ymin=0,
        ymax=100,
        ylabel={$\text{TPR}_{\text{user}} @ \text{FPR} = 0\%$},
        xlabel={Number of individuals},
    ]

    \addplot[name path=U, draw=none, mark=none]table[x=num, y expr=\thisrow{mean} + \thisrow{std}] {\liraC};
    \addplot[name path=L, draw=none, mark = none] table[x=num, y expr=\thisrow{mean} - \thisrow{std}] {\liraC};    
    \addplot+[fill=blue, fill opacity=0.20, draw=none, forget plot] fill between[of=U and L];
    \addplot+[color=blue, solid, mark=*] table[x=num, y=mean]{\liraC};

    \addplot[name path=U, draw=none, mark=none]table[x=num, y expr=\thisrow{mean} + \thisrow{std}] {\msliraC};
    \addplot[name path=L, draw=none, mark = none] table[x=num, y expr=\thisrow{mean} - \thisrow{std}] {\msliraC};
    \addplot[fill=red, fill opacity=0.20, draw=none, forget plot] fill between[of=U and L];
    \addplot+[color=red, solid, mark=*] table[x=num, y=mean]{\msliraC};

    \addplot[name path=U, draw=none, mark=none]table[x=num, y expr=\thisrow{mean} + \thisrow{std}] {\dtsC};
    \addplot[name path=L, draw=none, mark = none] table[x=num, y expr=\thisrow{mean} - \thisrow{std}] {\dtsC};
    \addplot+[fill=brown, fill opacity=0.20, draw=none, forget plot] fill between[of=U and L];
    \addplot+[color=brown, solid, mark=*] table[x=num, y=mean]{\dtsC};
    
    \end{groupplot}
  \node at ($(group c1r1.north east)!0.5!(group c3r1.north west)+ (0, 0.5cm)$) {\pgfplotslegendfromname{sharedlegend}};
\end{tikzpicture}

%% file: Results_includes/fig_access_ratio_EEG.tex
\pgfplotstableread{
num   mean    std    
0.00   0.274950  0.158620
0.05   0.581203  0.123559
0.10   1.000000  0.000000
0.15   1.000000  0.000000
0.20   1.000000  0.000000
0.25   1.000000  0.000000
0.30   1.000000  0.000000
0.35   1.000000  0.000000
0.40   1.000000  0.000000
0.45   1.000000  0.000000
0.50   1.000000  0.000000
0.55   1.000000  0.000000
0.60   1.000000  0.000000
0.65   1.000000  0.000000
0.70   1.000000  0.000000
0.75   1.000000  0.000000
0.80   1.000000  0.000000
0.85   1.000000  0.000000
0.90   1.000000  0.000000
0.95   1.000000  0.000000
1.00   1.000000  0.000000
}\liraA

\pgfplotstableread{
num   mean    std    
0.00   0.315950  0.169313
0.05   0.477852  0.114244
0.10   0.688792  0.017642
0.15   1.000000  0.000000
0.20   1.000000  0.000000
0.25   1.000000  0.000000
0.30   1.000000  0.000000
0.35   1.000000  0.000000
0.40   1.000000  0.000000
0.45   1.000000  0.000000
0.50   1.000000  0.000000
0.55   1.000000  0.000000
0.60   1.000000  0.000000
0.65   1.000000  0.000000
0.70   1.000000  0.000000
0.75   1.000000  0.000000
0.80   1.000000  0.000000
0.85   1.000000  0.000000
0.90   1.000000  0.000000
0.95   1.000000  0.000000
1.00   1.000000  0.000000
}\msliraA

\pgfplotstableread{
num   mean    std    
0.00   0.125850  0.179406
0.05   0.389600  0.020304
0.10   0.492700  0.017664
0.15   0.595550  0.014244
0.20   0.667750  0.010370
0.25   0.789050  0.006830
0.30   0.899500  0.004977
0.35   0.899800  0.003158
0.40   0.900000  0.000000
0.45   0.900000  0.000000
0.50   0.900000  0.000000
0.55   0.900000  0.000000
0.60   0.900000  0.000000
0.65   0.900000  0.000000
0.70   0.900000  0.000000
0.75   0.900000  0.000000
0.80   0.900000  0.000000
0.85   0.900000  0.000000
0.90   0.900000  0.000000
0.95   0.900000  0.000000
1.00   0.900000  0.000000
}\dtsA

\begin{tikzpicture}
\begin{axis}[
    width=7cm,
    height=7cm,
    xmin=0, xmax=1,
    ymin=0.2, ymax=1.05,
    grid=major,
        ylabel={$\text{TPR} @ \text{FPR} = 0\%$},
        xlabel={$n_{\mathrm{c}}/n$},
            label style={font=\Large},          
    tick label style={font=\Large},
    legend style={
        at={(0.1,0.15)},   
        anchor=west,
        legend columns=1,
        nodes={anchor=east, align=right, font= \normalsize}, %
    },
]

    \addplot[name path=U1, draw=none, mark=none, forget plot]
        table[x=num, y expr=\thisrow{mean} + \thisrow{std}] {\liraA};
    \addplot[name path=L1, draw=none, mark=none, forget plot]
        table[x=num, y expr=\thisrow{mean} - \thisrow{std}] {\liraA};
    \addplot[fill=blue, fill opacity=0.20, draw=none, forget plot]
        fill between[of=U1 and L1];
    \addplot+[color=blue, solid, mark=*,  line width=1.2pt]
        table[x=num, y=mean]{\liraA};
    \addlegendentry{LiRA-Online (TS2Vec)}

    \addplot[name path=U2, draw=none, mark=none, forget plot]
        table[x=num, y expr=\thisrow{mean} + \thisrow{std}] {\msliraA};
    \addplot[name path=L2, draw=none, mark=none, forget plot]
        table[x=num, y expr=\thisrow{mean} - \thisrow{std}] {\msliraA};
    \addplot[fill=red, fill opacity=0.20, draw=none, forget plot]
        fill between[of=U2 and L2];
    \addplot+[color=red, solid, mark=*,  line width=1.2pt]
        table[x=num, y=mean]{\msliraA};
    \addlegendentry{Multi-Signal LiRA-Online}

    \addplot[name path=U3, draw=none, mark=none, forget plot]
        table[x=num, y expr=\thisrow{mean} + \thisrow{std}] {\dtsA};
    \addplot[name path=L3, draw=none, mark=none, forget plot]
        table[x=num, y expr=\thisrow{mean} - \thisrow{std}] {\dtsA};
    \addplot[fill=brown, fill opacity=0.20, draw=none, forget plot]
        fill between[of=U3 and L3];
    \addplot+[color=brown, solid, mark=*,  line width=1.2pt]
        table[x=num, y=mean]{\dtsA};
    \addlegendentry{DTS (InceptionTime)}

\end{axis}
\end{tikzpicture}

%% file: 08.tex
\section{Conclusion}
In conclusion, we bring membership inference attacks to deep learning–based time series forecasting, adapting LiRA and RMIA, introducing a multi-signal LiRA, and proposing the deep learning–based DTS attack. Evaluations on real datasets (EEG and ELD) with LSTM and N-HiTS show that these models are vulnerable, particularly under user-level threat models, despite early stopping and good generalization. 
Online attacks substantially outperform offline ones, and that user-level attacks can reveal severe vulnerabilities missed by record-level evaluation. 
Ablations further show that longer forecasting horizons increase leakage, while larger populations reduce it. 
Our results establish strong baselines and highlight key factors influencing privacy risks in time series forecasting, informing the design of both future attacks and defenses.

While our study provides a starting point, it is limited to two datasets and two model architectures. Expanding evaluations to other domains (e.g., ECG signals, financial transactions, IoT sensors, GPS traces), alternative architectures such as transformers and temporal convolutional networks, and settings incorporating exogenous variables or probabilistic forecasts could yield a more complete picture of vulnerabilities. Another important limitation is the relatively small number of individuals in our datasets, which can introduce instability in the results; extending evaluations to larger cohorts would help assess the robustness and generalizability of the findings. Moreover, refining user-level inference, moving beyond independent score aggregation to end-to-end approaches, remains an important open challenge.
Addressing these directions is key to understanding privacy risks in time series forecasting.

\subsection*{Ethics Statement}
This work aims to improve risk assessment for sensitive time series data by identifying vulnerabilities in forecasting models. 
This work relies on public datasets and the code is open-sourced to support reproducible research.

%% file: 09.tex
\onecolumn
\appendix

\subsection{ROC Curves} 

\label{sec:appendix_roc}
In this appendix, we report the ROC curves corresponding to the results presented in Sec.~\ref{sec:experimental_results} for both the online and the offline setting.

\input{attack_ROC_online_fig}

\input{attack_ROC_offline_fig}

\onecolumn

\subsection{Mitigation via DP-SGD}
\label{sec:DP-SGD}

Membership inference is fundamentally tied to memorization, leading to systematic differences between predictions on members and non-members~\cite{carlini2022first, carlini2019secret}. 
Some mitigation strategies therefore aim to reduce such memorization, for example through regularization techniques such as dropout, weight decay, and early stopping. In all main experiments in Section~\ref{sec:experimental_results}, we employ early stopping as part of the standard training procedure.

Beyond regularization-based methods, differential privacy is commonly adopted to provide further resilience. 
In particular, DP-SGD~\cite{abadi2016deep} injects calibrated noise during training and offers formal privacy guarantees. 

To examine this effect in forecasting models, we conduct an experiment where an N-HiTS target model is trained on the ELD dataset using DP-SGD with the clipping norm fixed at $C = 10$, batch size $1024$, and the noise multiplier varied as $\sigma \in \lbrace0.0, 0.2,0.8,1.0\rbrace$. 
We then attack the target model under the record-level threat model using DTS and Multi-Signal LiRA in the online setting.

Table~\ref{tab:dpsgd} reports attack performance and target-model losses (MSE, MAE, SMAPE, ND) across different noise levels.
Results without DP-SGD ($\sigma= 0$ and $C= \text{None}$) are also included.
As expected, more noise consistently weakens the MIA while deteriorating the losses. 
In particular, the performance of the DTS attack under the most private setting ($\sigma=1.0$) yields TPR of $3.26\%$ at an FPR of $0.1\%$.
This is less than half the TPR observed without DP-SGD.

\begin{table*}[tb]
\captionof{table}{Target-model losses for N-HiTS trained on ELD under DP-SGD with clipping $C = 10$, batch size $1024$, and noise multipliers $\sigma \in \{0, 0.2, 0.8, 1\}$, along with DTS-online and Multi-Signal LiRA-Online attack performance on the mentioned models at $0.1\%$ FPR. 
The model losses and attack performance for the non–DP-SGD baseline ($\sigma = 0 $, and $C=\text{None}$) are also reported for reference.}
\label{tab:dpsgd}
\centering
\resizebox{\linewidth}{!}{
\begin{tabular}{cc|cc|cccc}
\multirow{2}{*}{\begin{tabular}{l} $\sigma$ \end{tabular}} &
\multirow{2}{*}{\begin{tabular}{l} $C$ \end{tabular}} &
\multirow{2}{*}{\begin{tabular}{c} DTS (InceptionTime) \\ $0.1\%$ FPR \end{tabular}} &
\multirow{2}{*}{\begin{tabular}{c} Multi-Signal LiRA \\ $0.1\%$ FPR \end{tabular}} &
\multicolumn{4}{c}{Losses} \\
  &   &    &                  & MSE  & MAE  & SMAPE  & ND \\ \hline
0.0       &  None    &     7.12 &   3.32 &    0.034314  &  0.088645    &   0.166489      & 0.163490   \\
0.0       &   10     &  5.69  &   3.02 &    0.344160  &   0.091787   &     0.178334    & 0.177658   \\
0.2       & 10       &  5.71 & 2.85 &  0.034814       &     0.094275  &   0.175760  &   0.173873      \\
0.8       & 10       & 4.91  & 1.91   &     0.037147  &  0.098007    &    0.181079  &    0.180756      \\
1.0       & 10       &  3.26  &   1.88   &   0.039224   &   0.105430   &   0.195587      &   0.195476
\end{tabular}
}
\end{table*}

%% file: attack_ROC_online_fig.tex
\pgfplotscreateplotcyclelist{mycolors}{
    {blue, mark=none},
    {red, mark=none},
    {green!60!black, mark=none},
    {orange, mark=none},
    {purple, mark=none},
    {brown, mark=none},
    {cyan!60!black, mark=none},
}

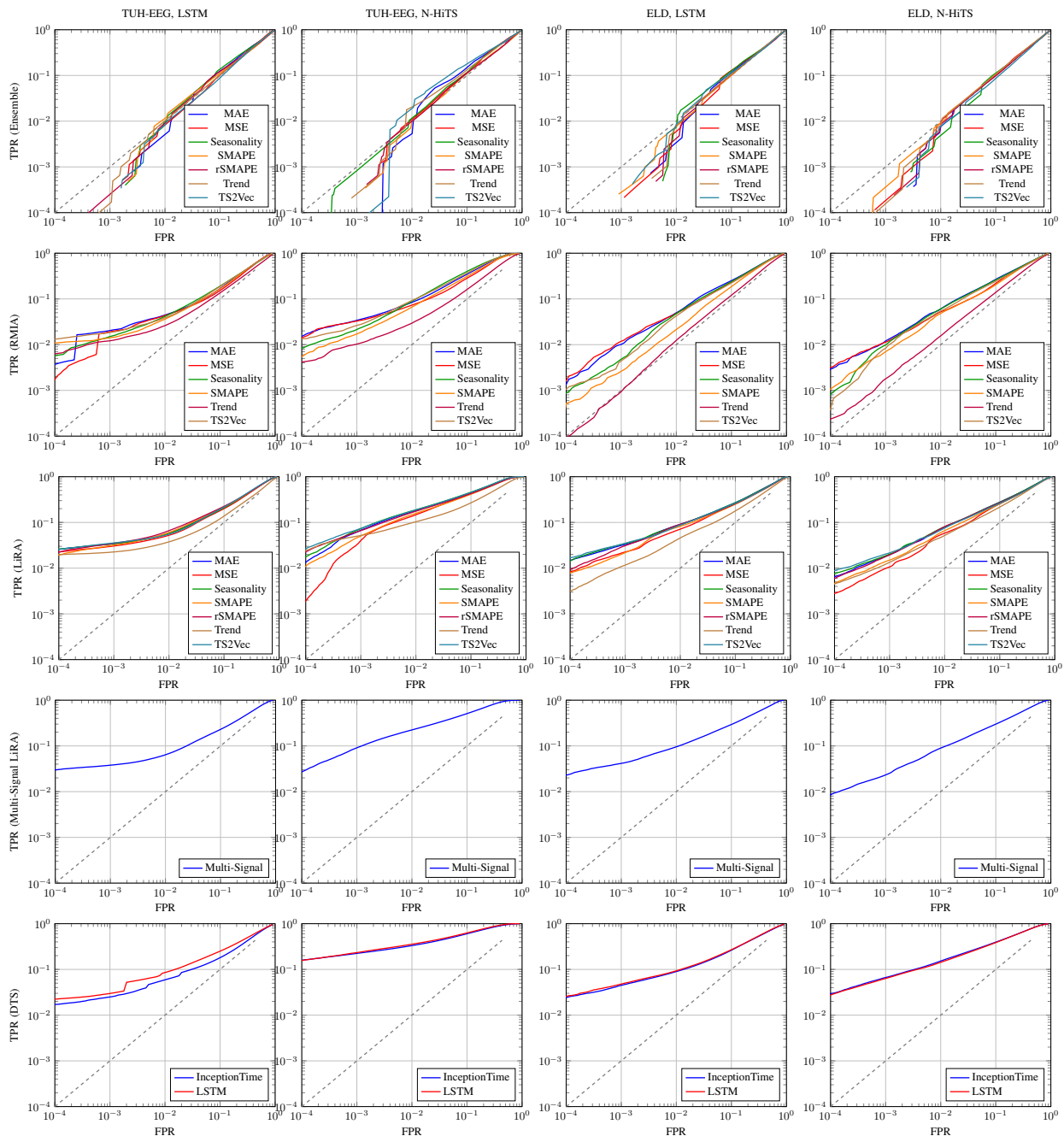
\begin{figure*}[ht]
\centering

\input{online_ROCs_rows/Ensemble}

\input{online_ROCs_rows/RMIA}

\input{online_ROCs_rows/LiRA}

 \input{online_ROCs_rows/Multi-LiRA}

 \input{online_ROCs_rows/DTS}

\par\vspace{1em}

\caption{ROC curves with logarithmic scales (TPR on the y-axis and FPR on the x-axis) for five membership inference attacks: Ensemble, RMIA, LiRA, Multi-Signal LiRA, and DTS, arranged in rows. Columns correspond to dataset–architecture pairs: EEG–LSTM, EEG–N-HiTS, ELD–LSTM, and ELD–N-HiTS. All results are for the \textbf{online} setting, with the Ensemble attack evaluated in audit mode. Within each subplot, curves correspond to individual attack signals for LiRA, RMIA, and Ensemble, all combined signals for Multi-Signal LiRA, and different classifier architectures for DTS. Each curve represents the mean over five independent runs.}
\label{fig:roc_online}
\end{figure*}

%% file: online_ROCs_rows/Ensemble.tex
\pgfplotsset{
  every axis/.append style={
    label style={font=\normalsize},
    tick label style={font=\normalsize},
    legend style={font=\normalsize},
    legend pos={south east},
  }
}

\begin{subfigure}[t]{0.22\textwidth}
\centering
\begin{tikzpicture}[scale=0.5]
  \begin{axis}[
    xlabel={FPR}, ylabel={TPR (Ensemble)},
    title={TUH-EEG, LSTM},
    each nth point={5},    
    xmode=log,
    ymode=log,
    xmin=1e-4, xmax=1,
    ymin=1e-4, ymax=1,
    grid=major,
    every axis plot/.append style={line width=1pt},
    cycle list name=mycolors
  ]
    
    \addplot [gray, dashed, domain=1e-4:1,samples=200, forget plot] {x};

    \addplot+[
    mark=none]
    table[x=fpr, y=tpr, col sep=comma] {data/EEG/Ensemble/LSTM-audit_mode/total/roc_MAELoss.csv};
    \addlegendentry{MAE}
    
    \addplot+[mark=none]
      table[x=fpr, y=tpr, col sep=comma] {data/EEG/Ensemble/LSTM-audit_mode/total/roc_MSELoss.csv};
    \addlegendentry{MSE}
    
    \addplot+[mark=none]
      table[x=fpr, y=tpr, col sep=comma] {data/EEG/Ensemble/LSTM-audit_mode/total/roc_SeasonalityLoss.csv};
    \addlegendentry{Seasonality}
    
    \addplot+[mark=none]
      table[x=fpr, y=tpr, col sep=comma] {data/EEG/Ensemble/LSTM-audit_mode/total/roc_SMAPELoss.csv};
    \addlegendentry{SMAPE}
    
    \addplot+[mark=none]
      table[x=fpr, y=tpr, col sep=comma] {data/EEG/Ensemble/LSTM-audit_mode/total/roc_RescaledSMAPELoss.csv};
    \addlegendentry{rSMAPE}
    
    \addplot+[mark=none]
      table[x=fpr, y=tpr, col sep=comma] {data/EEG/Ensemble/LSTM-audit_mode/total/roc_TrendLoss.csv};
    \addlegendentry{Trend}
    
    \addplot+[mark=none]
      table[x=fpr, y=tpr, col sep=comma] {data/EEG/Ensemble/LSTM-audit_mode/total/roc_TS2VecLoss.csv};
    \addlegendentry{TS2Vec}
    
  \end{axis}
\end{tikzpicture}
\end{subfigure}
\begin{subfigure}[t]{0.22\textwidth}
\centering
\begin{tikzpicture}[scale=0.5]
  \begin{axis}[
    xlabel={FPR}, ylabel={},
    title={TUH-EEG, N-HiTS},
    each nth point={5},    
    xmode=log,
    ymode=log,
    xmin=1e-4, xmax=1,
    ymin=1e-4, ymax=1,
    grid=major,
    every axis plot/.append style={line width=1pt},
    cycle list name=mycolors
  ]
  \addplot [gray, dashed, domain=1e-4:1, samples=200, forget plot] {x};
  
    \addplot+[mark=none]
      table[x=fpr, y=tpr, col sep=comma] {data/EEG/Ensemble/NHiTS-audit_mode/total/roc_MAELoss.csv};
    \addlegendentry{MAE}
    
    \addplot+[mark=none]
      table[x=fpr, y=tpr, col sep=comma] {data/EEG/Ensemble/NHiTS-audit_mode/total/roc_MSELoss.csv};
    \addlegendentry{MSE}
    
    \addplot+[mark=none]
      table[x=fpr, y=tpr, col sep=comma] {data/EEG/Ensemble/NHiTS-audit_mode/total/roc_SeasonalityLoss.csv};
    \addlegendentry{Seasonality}
    
    \addplot+[mark=none]
      table[x=fpr, y=tpr, col sep=comma] {data/EEG/Ensemble/NHiTS-audit_mode/total/roc_SMAPELoss.csv};
    \addlegendentry{SMAPE}
    
    \addplot+[mark=none]
      table[x=fpr, y=tpr, col sep=comma] {data/EEG/Ensemble/NHiTS-audit_mode/total/roc_RescaledSMAPELoss.csv};
    \addlegendentry{rSMAPE}
    
    \addplot+[mark=none]
      table[x=fpr, y=tpr, col sep=comma] {data/EEG/Ensemble/NHiTS-audit_mode/total/roc_TrendLoss.csv};
    \addlegendentry{Trend}
    
    \addplot+[mark=none]
      table[x=fpr, y=tpr, col sep=comma] {data/EEG/Ensemble/NHiTS-audit_mode/total/roc_TS2VecLoss.csv};
    \addlegendentry{TS2Vec}
  \end{axis}
\end{tikzpicture}
\end{subfigure}
\begin{subfigure}[t]{0.22\textwidth}
\centering
\begin{tikzpicture}[scale=0.5]
  \begin{axis}[
    xlabel={FPR}, ylabel={},
    title={ELD, LSTM},
        each nth point={5},    
    xmode=log,
    ymode=log,
    xmin=1e-4, xmax=1,
    ymin=1e-4, ymax=1,
    grid=major,
    every axis plot/.append style={line width=1pt},
    cycle list name=mycolors
  ]
  \addplot [gray, dashed, domain=1e-4:1, samples=200, forget plot] {x};
  
    \addplot+[mark=none]
      table[x=fpr, y=tpr, col sep=comma] {data/ELD/Ensemble/LSTM-audit_mode/total/roc_MAELoss.csv};
    \addlegendentry{MAE}
    
    \addplot+[mark=none]
      table[x=fpr, y=tpr, col sep=comma] {data/ELD/Ensemble/LSTM-audit_mode/total/roc_MSELoss.csv};
    \addlegendentry{MSE}
    
    \addplot+[mark=none]
      table[x=fpr, y=tpr, col sep=comma] {data/ELD/Ensemble/LSTM-audit_mode/total/roc_SeasonalityLoss.csv};
    \addlegendentry{Seasonality}
    
    \addplot+[mark=none]
      table[x=fpr, y=tpr, col sep=comma] {data/ELD/Ensemble/LSTM-audit_mode/total/roc_SMAPELoss.csv};
    \addlegendentry{SMAPE}
    
    \addplot+[mark=none]
      table[x=fpr, y=tpr, col sep=comma] {data/ELD/Ensemble/LSTM-audit_mode/total/roc_RescaledSMAPELoss.csv};
    \addlegendentry{rSMAPE}
    
    \addplot+[mark=none]
      table[x=fpr, y=tpr, col sep=comma] {data/ELD/Ensemble/LSTM-audit_mode/total/roc_TrendLoss.csv};
    \addlegendentry{Trend}
    
    \addplot+[mark=none]
      table[x=fpr, y=tpr, col sep=comma] {data/ELD/Ensemble/LSTM-audit_mode/total/roc_TS2VecLoss.csv};
    \addlegendentry{TS2Vec}
  \end{axis}
\end{tikzpicture}
\end{subfigure}
\begin{subfigure}[t]{0.22\textwidth}
\centering
\begin{tikzpicture}[scale=0.5]
  \begin{axis}[
    xlabel={FPR}, ylabel={},
    title={ELD, N-HiTS},
        each nth point={5},    
    xmode=log,
    ymode=log,
    xmin=1e-4, xmax=1,
    ymin=1e-4, ymax=1,
    grid=major,
    every axis plot/.append style={line width=1pt},
    cycle list name=mycolors
  ]
  \addplot [gray, dashed, domain=1e-4:1, samples=200, forget plot] {x};
  
    \addplot+[mark=none]
      table[x=fpr, y=tpr, col sep=comma] {data/ELD/Ensemble/NHiTS-audit_mode/total/roc_MAELoss.csv};
    \addlegendentry{MAE}
    
    \addplot+[mark=none]
      table[x=fpr, y=tpr, col sep=comma] {data/ELD/Ensemble/NHiTS-audit_mode/total/roc_MSELoss.csv};
    \addlegendentry{MSE}
    
    \addplot+[mark=none]
      table[x=fpr, y=tpr, col sep=comma] {data/ELD/Ensemble/NHiTS-audit_mode/total/roc_SeasonalityLoss.csv};
    \addlegendentry{Seasonality}
    
    \addplot+[mark=none]
      table[x=fpr, y=tpr, col sep=comma] {data/ELD/Ensemble/NHiTS-audit_mode/total/roc_SMAPELoss.csv};
    \addlegendentry{SMAPE}
    
    \addplot+[mark=none]
      table[x=fpr, y=tpr, col sep=comma] {data/ELD/Ensemble/NHiTS-audit_mode/total/roc_RescaledSMAPELoss.csv};
    \addlegendentry{rSMAPE}
    
    \addplot+[mark=none]
      table[x=fpr, y=tpr, col sep=comma] {data/ELD/Ensemble/NHiTS-audit_mode/total/roc_TrendLoss.csv};
    \addlegendentry{Trend}
    
    \addplot+[mark=none]
      table[x=fpr, y=tpr, col sep=comma] {data/ELD/Ensemble/NHiTS-audit_mode/total/roc_TS2VecLoss.csv};
    \addlegendentry{TS2Vec}
  \end{axis}
\end{tikzpicture}
\end{subfigure}

%% file: online_ROCs_rows/RMIA.tex
\pgfplotsset{
  every axis/.append style={
    label style={font=\normalsize},
    tick label style={font=\normalsize},
    legend style={font=\normalsize},
  },
}

\begin{subfigure}[t]{0.22\textwidth}
\centering
\begin{tikzpicture}[scale=0.5]
  \begin{axis}[
    xmode=log, ymode=log,
    log basis y = 10,
    each nth point={10},
    xmin=1e-4, ymin=1e-4, xmax=1, ymax=1,
    xlabel={FPR}, ylabel={TPR (RMIA)},
    grid=major,
    legend pos={south east},
    legend style={/tikz/every node/.append style={anchor=west}},
    every axis plot/.append style={line width=1pt},
    cycle list name=mycolors,
  ]
    
    \addplot [gray, dashed, domain=1e-4:1, samples=100, forget plot] {x};

    \addplot+[
    mark=none]
    table[x=fpr, y=tpr, col sep=comma] {data/EEG/RMIA/LSTM-online/total/roc_MAELoss.csv};
    \addlegendentry{MAE}
    
    \addplot+[mark=none]
      table[x=fpr, y=tpr, col sep=comma] {data/EEG/RMIA/LSTM-online/total/roc_MSELoss.csv};
    \addlegendentry{MSE}
    
    \addplot+[mark=none]
      table[x=fpr, y=tpr, col sep=comma] {data/EEG/RMIA/LSTM-online/total/roc_SeasonalityLoss.csv};
    \addlegendentry{Seasonality}
    
    \addplot+[mark=none]
      table[x=fpr, y=tpr, col sep=comma] {data/EEG/RMIA/LSTM-online/total/roc_SMAPELoss.csv};
    \addlegendentry{SMAPE}
    
    \addplot+[mark=none]
      table[x=fpr, y=tpr, col sep=comma] {data/EEG/RMIA/LSTM-online/total/roc_TrendLoss.csv};
    \addlegendentry{Trend}
    
    \addplot+[mark=none]
      table[x=fpr, y=tpr, col sep=comma] {data/EEG/RMIA/LSTM-online/total/roc_TS2VecLoss.csv};
    \addlegendentry{TS2Vec}
    
  \end{axis}
\end{tikzpicture}
\end{subfigure}
\begin{subfigure}[t]{0.22\textwidth}
\centering
\begin{tikzpicture}[scale=0.5]
  \begin{axis}[
    xmode=log, ymode=log,
    xmin=1e-4, ymin=1e-4, xmax=1, ymax=1,
    xlabel={FPR}, ylabel={},
    grid=major,
    each nth point={10},
    legend pos={south east},
    legend style={/tikz/every node/.append style={anchor=west}},
    every axis plot/.append style={line width=1pt},
    cycle list name=mycolors,
  ]
  \addplot [gray, dashed, domain=1e-4:1, samples=100, forget plot] {x};
  
    \addplot+[mark=none]
      table[x=fpr, y=tpr, col sep=comma] {data/EEG/RMIA/NHiTS-online/total/roc_MAELoss.csv};
    \addlegendentry{MAE}
    
    \addplot+[mark=none]
      table[x=fpr, y=tpr, col sep=comma] {data/EEG/RMIA/NHiTS-online/total/roc_MSELoss.csv};
    \addlegendentry{MSE}
    
    \addplot+[mark=none]
      table[x=fpr, y=tpr, col sep=comma] {data/EEG/RMIA/NHiTS-online/total/roc_SeasonalityLoss.csv};
    \addlegendentry{Seasonality}
    
    \addplot+[mark=none]
      table[x=fpr, y=tpr, col sep=comma] {data/EEG/RMIA/NHiTS-online/total/roc_SMAPELoss.csv};
    \addlegendentry{SMAPE}
    
    \addplot+[mark=none]
      table[x=fpr, y=tpr, col sep=comma] {data/EEG/RMIA/NHiTS-online/total/roc_TrendLoss.csv};
    \addlegendentry{Trend}
    
    \addplot+[mark=none]
      table[x=fpr, y=tpr, col sep=comma] {data/EEG/RMIA/NHiTS-online/total/roc_TS2VecLoss.csv};
    \addlegendentry{TS2Vec}
  \end{axis}
\end{tikzpicture}
\end{subfigure}
\begin{subfigure}[t]{0.22\textwidth}
\centering
\begin{tikzpicture}[scale=0.5]
  \begin{axis}[
    xmode=log, ymode=log,
    xmin=1e-4, ymin=1e-4, xmax=1, ymax=1,
    xlabel={FPR}, ylabel={},
    grid=major,
    each nth point={10},
    legend pos={south east},
    legend style={/tikz/every node/.append style={anchor=west}},
    every axis plot/.append style={line width=1pt},
    cycle list name=mycolors,
  ]
  \addplot [gray, dashed, domain=1e-4:1, samples=100, forget plot] {x};
  
    \addplot+[mark=none]
      table[x=fpr, y=tpr, col sep=comma] {data/ELD/RMIA/LSTM-online/total/roc_MAELoss.csv};
    \addlegendentry{MAE}
    
    \addplot+[mark=none]
      table[x=fpr, y=tpr, col sep=comma] {data/ELD/RMIA/LSTM-online/total/roc_MSELoss.csv};
    \addlegendentry{MSE}
    
    \addplot+[mark=none]
      table[x=fpr, y=tpr, col sep=comma] {data/ELD/RMIA/LSTM-online/total/roc_SeasonalityLoss.csv};
    \addlegendentry{Seasonality}
    
    \addplot+[mark=none]
      table[x=fpr, y=tpr, col sep=comma] {data/ELD/RMIA/LSTM-online/total/roc_SMAPELoss.csv};
    \addlegendentry{SMAPE}
    
    \addplot+[mark=none]
      table[x=fpr, y=tpr, col sep=comma] {data/ELD/RMIA/LSTM-online/total/roc_TrendLoss.csv};
    \addlegendentry{Trend}
    
    \addplot+[mark=none]
      table[x=fpr, y=tpr, col sep=comma] {data/ELD/RMIA/LSTM-online/total/roc_TS2VecLoss.csv};
    \addlegendentry{TS2Vec}
  \end{axis}
\end{tikzpicture}
\end{subfigure}
\begin{subfigure}[t]{0.22\textwidth}
\centering
\begin{tikzpicture}[scale=0.5]
  \begin{axis}[
    xmode=log, ymode=log,
    xmin=1e-4, ymin=1e-4, xmax=1, ymax=1,
    xlabel={FPR}, ylabel={},
    grid=major,
    each nth point={10},
    legend pos={south east},
    legend style={/tikz/every node/.append style={anchor=west}},
    every axis plot/.append style={line width=1pt},
    cycle list name=mycolors,
  ]
  \addplot [gray, dashed, domain=1e-4:1, samples=100, forget plot] {x};
  
    \addplot+[mark=none]
      table[x=fpr, y=tpr, col sep=comma] {data/ELD/RMIA/NHiTS-online/total/roc_MAELoss.csv};
    \addlegendentry{MAE}
    
    \addplot+[mark=none]
      table[x=fpr, y=tpr, col sep=comma] {data/ELD/RMIA/NHiTS-online/total/roc_MSELoss.csv};
    \addlegendentry{MSE}
    
    \addplot+[mark=none]
      table[x=fpr, y=tpr, col sep=comma] {data/ELD/RMIA/NHiTS-online/total/roc_SeasonalityLoss.csv};
    \addlegendentry{Seasonality}
    
    \addplot+[mark=none]
      table[x=fpr, y=tpr, col sep=comma] {data/ELD/RMIA/NHiTS-online/total/roc_SMAPELoss.csv};
    \addlegendentry{SMAPE}
    
    \addplot+[mark=none]
      table[x=fpr, y=tpr, col sep=comma] {data/ELD/RMIA/NHiTS-online/total/roc_TrendLoss.csv};
    \addlegendentry{Trend}
    
    \addplot+[mark=none]
      table[x=fpr, y=tpr, col sep=comma] {data/ELD/RMIA/NHiTS-online/total/roc_TS2VecLoss.csv};
    \addlegendentry{TS2Vec}
  \end{axis}
\end{tikzpicture}
\end{subfigure}

%% file: online_ROCs_rows/LiRA.tex
\pgfplotsset{
  every axis/.append style={
    label style={font=\normalsize},
    tick label style={font=\normalsize},
    legend style={font=\normalsize},
  }
}

\begin{subfigure}[t]{0.22\textwidth}
\centering
\begin{tikzpicture}[scale=0.5]
  \begin{axis}[
    xmode=log, ymode=log,
    log basis y = 10,
    each nth point={10},
    xmin=1e-4, ymin=1e-4, xmax=1, ymax=1,
    xlabel={FPR}, ylabel={TPR (LiRA)},
    title={},
    grid=major,
    legend pos={south east},
    legend style={/tikz/every node/.append style={anchor=west}},
    every axis plot/.append style={line width=1pt},
    cycle list name=mycolors,
  ]    
    \addplot [gray, dashed, domain=1e-4:1, samples=100, forget plot] {x};

    \addplot+[
    mark=none]
    table[x=fpr, y=tpr, col sep=comma] {data/EEG/LiRA/LSTM-online/total/roc_MAELoss.csv};
    \addlegendentry{MAE}
    
    \addplot+[mark=none]
      table[x=fpr, y=tpr, col sep=comma] {data/EEG/LiRA/LSTM-online/total/roc_MSELoss.csv};
    \addlegendentry{MSE}
    
    \addplot+[mark=none]
      table[x=fpr, y=tpr, col sep=comma] {data/EEG/LiRA/LSTM-online/total/roc_SeasonalityLoss.csv};
    \addlegendentry{Seasonality}
    
    \addplot+[mark=none]
      table[x=fpr, y=tpr, col sep=comma] {data/EEG/LiRA/LSTM-online/total/roc_SMAPELoss.csv};
    \addlegendentry{SMAPE}
    
    \addplot+[mark=none]
      table[x=fpr, y=tpr, col sep=comma] {data/EEG/LiRA/LSTM-online/total/roc_RescaledSMAPELoss.csv};
    \addlegendentry{rSMAPE}
    
    \addplot+[mark=none]
      table[x=fpr, y=tpr, col sep=comma] {data/EEG/LiRA/LSTM-online/total/roc_TrendLoss.csv};
    \addlegendentry{Trend}
    
    \addplot+[mark=none]
      table[x=fpr, y=tpr, col sep=comma] {data/EEG/LiRA/LSTM-online/total/roc_TS2VecLoss.csv};
    \addlegendentry{TS2Vec}
    
  \end{axis}
\end{tikzpicture}
\end{subfigure}
\begin{subfigure}[t]{0.22\textwidth}
\centering
\begin{tikzpicture}[scale=0.5]
  \begin{axis}[
    xmode=log, ymode=log,
    xmin=1e-4, ymin=1e-4, xmax=1, ymax=1,
    xlabel={FPR}, ylabel={},
    grid=major,
    each nth point={10},
    legend pos={south east},
    legend style={/tikz/every node/.append style={anchor=west}},
    every axis plot/.append style={line width=1pt},
    cycle list name=mycolors,
  ]
  \addplot [gray, dashed, domain=1e-4:1, samples=100, forget plot] {x};
  
    \addplot+[mark=none]
      table[x=fpr, y=tpr, col sep=comma] {data/EEG/LiRA/NHiTS-online/total/roc_MAELoss.csv};
    \addlegendentry{MAE}
    
    \addplot+[mark=none]
      table[x=fpr, y=tpr, col sep=comma] {data/EEG/LiRA/NHiTS-online/total/roc_MSELoss.csv};
    \addlegendentry{MSE}
    
    \addplot+[mark=none]
      table[x=fpr, y=tpr, col sep=comma] {data/EEG/LiRA/NHiTS-online/total/roc_SeasonalityLoss.csv};
    \addlegendentry{Seasonality}
    
    \addplot+[mark=none]
      table[x=fpr, y=tpr, col sep=comma] {data/EEG/LiRA/NHiTS-online/total/roc_SMAPELoss.csv};
    \addlegendentry{SMAPE}
    
    \addplot+[mark=none]
      table[x=fpr, y=tpr, col sep=comma] {data/EEG/LiRA/NHiTS-online/total/roc_RescaledSMAPELoss.csv};
    \addlegendentry{rSMAPE}
    
    \addplot+[mark=none]
      table[x=fpr, y=tpr, col sep=comma] {data/EEG/LiRA/NHiTS-online/total/roc_TrendLoss.csv};
    \addlegendentry{Trend}
    
    \addplot+[mark=none]
      table[x=fpr, y=tpr, col sep=comma] {data/EEG/LiRA/NHiTS-online/total/roc_TS2VecLoss.csv};
    \addlegendentry{TS2Vec}
  \end{axis}
\end{tikzpicture}
\end{subfigure}
\begin{subfigure}[t]{0.22\textwidth}
\centering
\begin{tikzpicture}[scale=0.5]
  \begin{axis}[
    xmode=log, ymode=log,
    xmin=1e-4, ymin=1e-4, xmax=1, ymax=1,
    xlabel={FPR}, ylabel={},
    grid=major,
    each nth point={10},
    legend pos={south east},
    legend style={/tikz/every node/.append style={anchor=west}},
    every axis plot/.append style={line width=1pt},
    cycle list name=mycolors,
  ]
  \addplot [gray, dashed, domain=1e-4:1, samples=100, forget plot] {x};
  
    \addplot+[mark=none]
      table[x=fpr, y=tpr, col sep=comma] {data/ELD/LiRA/LSTM-online/total/roc_MAELoss.csv};
    \addlegendentry{MAE}
    
    \addplot+[mark=none]
      table[x=fpr, y=tpr, col sep=comma] {data/ELD/LiRA/LSTM-online/total/roc_MSELoss.csv};
    \addlegendentry{MSE}
    
    \addplot+[mark=none]
      table[x=fpr, y=tpr, col sep=comma] {data/ELD/LiRA/LSTM-online/total/roc_SeasonalityLoss.csv};
    \addlegendentry{Seasonality}
    
    \addplot+[mark=none]
      table[x=fpr, y=tpr, col sep=comma] {data/ELD/LiRA/LSTM-online/total/roc_SMAPELoss.csv};
    \addlegendentry{SMAPE}
    
    \addplot+[mark=none]
      table[x=fpr, y=tpr, col sep=comma] {data/ELD/LiRA/LSTM-online/total/roc_RescaledSMAPELoss.csv};
    \addlegendentry{rSMAPE}
    
    \addplot+[mark=none]
      table[x=fpr, y=tpr, col sep=comma] {data/ELD/LiRA/LSTM-online/total/roc_TrendLoss.csv};
    \addlegendentry{Trend}
    
    \addplot+[mark=none]
      table[x=fpr, y=tpr, col sep=comma] {data/ELD/LiRA/LSTM-online/total/roc_TS2VecLoss.csv};
    \addlegendentry{TS2Vec}
  \end{axis}
\end{tikzpicture}
\end{subfigure}
\begin{subfigure}[t]{0.22\textwidth}
\centering
\begin{tikzpicture}[scale=0.5]
  \begin{axis}[
    xmode=log, ymode=log,
    xmin=1e-4, ymin=1e-4, xmax=1, ymax=1,
    xlabel={FPR}, ylabel={},
    grid=major,
    each nth point={10},
    legend pos={south east},
    legend style={/tikz/every node/.append style={anchor=west}},
    every axis plot/.append style={line width=1pt},
    cycle list name=mycolors,
  ]
  \addplot [gray, dashed, domain=1e-4:1, samples=100, forget plot] {x};
  
    \addplot+[mark=none]
      table[x=fpr, y=tpr, col sep=comma] {data/ELD/LiRA/NHiTS-online/total/roc_MAELoss.csv};
    \addlegendentry{MAE}
    
    \addplot+[mark=none]
      table[x=fpr, y=tpr, col sep=comma] {data/ELD/LiRA/NHiTS-online/total/roc_MSELoss.csv};
    \addlegendentry{MSE}
    
    \addplot+[mark=none]
      table[x=fpr, y=tpr, col sep=comma] {data/ELD/LiRA/NHiTS-online/total/roc_SeasonalityLoss.csv};
    \addlegendentry{Seasonality}
    
    \addplot+[mark=none]
      table[x=fpr, y=tpr, col sep=comma] {data/ELD/LiRA/NHiTS-online/total/roc_SMAPELoss.csv};
    \addlegendentry{SMAPE}
    
    \addplot+[mark=none]
      table[x=fpr, y=tpr, col sep=comma] {data/ELD/LiRA/NHiTS-online/total/roc_RescaledSMAPELoss.csv};
    \addlegendentry{rSMAPE}
    
    \addplot+[mark=none]
      table[x=fpr, y=tpr, col sep=comma] {data/ELD/LiRA/NHiTS-online/total/roc_TrendLoss.csv};
    \addlegendentry{Trend}
    
    \addplot+[mark=none]
      table[x=fpr, y=tpr, col sep=comma] {data/ELD/LiRA/NHiTS-online/total/roc_TS2VecLoss.csv};
    \addlegendentry{TS2Vec}
  \end{axis}
\end{tikzpicture}
\end{subfigure}

%% file: online_ROCs_rows/Multi-LiRA.tex
\pgfplotsset{
  every axis/.append style={
    label style={font=\normalsize},
    tick label style={font=\normalsize},
    legend style={font=\normalsize},
  }
}

\begin{subfigure}[t]{0.22\textwidth}
\centering
\begin{tikzpicture}[scale=0.5]
  \begin{axis}[
    xmode=log, ymode=log,
    log basis y = 10,
    each nth point={10},
    xmin=1e-4, ymin=1e-4, xmax=1, ymax=1,
    xlabel={FPR}, ylabel={TPR (Multi-Signal LiRA)},
    grid=major,
    legend pos={south east},
    legend style={/tikz/every node/.append style={anchor=west}},
    every axis plot/.append style={line width=1pt},
    cycle list name=mycolors,
  ]

    \addplot [gray, dashed, domain=1e-4:1, samples=100, forget plot] {x};

    \addplot+[mark=none]
      table[x=fpr, y=tpr, col sep=comma] {data/EEG/Multi-LiRA/LSTM-online/total/roc.csv};
    \addlegendentry{Multi-Signal}
    
  \end{axis}
\end{tikzpicture}
\end{subfigure}
\begin{subfigure}[t]{0.22\textwidth}
\centering
\begin{tikzpicture}[scale=0.5]
  \begin{axis}[
    xmode=log, ymode=log,
    xmin=1e-4, ymin=1e-4, xmax=1, ymax=1,
    xlabel={FPR}, ylabel={},
    grid=major,
    each nth point={10},
    legend pos={south east},
    legend style={/tikz/every node/.append style={anchor=west}},
    every axis plot/.append style={line width=1pt},
    cycle list name=mycolors,
  ]
  \addplot [gray, dashed, domain=1e-4:1, samples=100, forget plot] {x};
  
    \addplot+[mark=none]
      table[x=fpr, y=tpr, col sep=comma] {data/EEG/Multi-LiRA/NHiTS-online/total/roc.csv};
    \addlegendentry{Multi-Signal}
    
  \end{axis}
\end{tikzpicture}
\end{subfigure}
\begin{subfigure}[t]{0.22\textwidth}
\centering
\begin{tikzpicture}[scale=0.5]
  \begin{axis}[
    xmode=log, ymode=log,
    xmin=1e-4, ymin=1e-4, xmax=1, ymax=1,
    xlabel={FPR}, ylabel={},
    grid=major,
    each nth point={10},
    legend pos={south east},
    legend style={/tikz/every node/.append style={anchor=west}},
    every axis plot/.append style={line width=1pt},
    cycle list name=mycolors,
  ]
  \addplot [gray, dashed, domain=1e-4:1, samples=100, forget plot] {x};
  
    \addplot+[mark=none]
      table[x=fpr, y=tpr, col sep=comma] {data/ELD/Multi-LiRA/LSTM-online/total/roc.csv};
    \addlegendentry{Multi-Signal}
  \end{axis}
\end{tikzpicture}
\end{subfigure}
\begin{subfigure}[t]{0.22\textwidth}
\centering
\begin{tikzpicture}[scale=0.5]
  \begin{axis}[
    xmode=log, ymode=log,
    xmin=1e-4, ymin=1e-4, xmax=1, ymax=1,
    xlabel={FPR}, ylabel={},
    grid=major,
    each nth point={10},
    legend pos={south east},
    legend style={/tikz/every node/.append style={anchor=west}},
    every axis plot/.append style={line width=1pt},
    cycle list name=mycolors,
  ]
  \addplot [gray, dashed, domain=1e-4:1, samples=100, forget plot] {x};
  
    \addplot+[mark=none]
      table[x=fpr, y=tpr, col sep=comma] {data/ELD/Multi-LiRA/NHiTS-online/total/roc.csv};
    \addlegendentry{Multi-Signal}
    
  \end{axis}
\end{tikzpicture}
\end{subfigure}

%% file: online_ROCs_rows/DTS.tex
\pgfplotsset{
  every axis/.append style={
    label style={font=\normalsize},
    tick label style={font=\normalsize},
    legend style={font=\normalsize},
  }
}

\begin{subfigure}[t]{0.22\textwidth}
\centering
\begin{tikzpicture}[scale=0.5]
  \begin{axis}[
    xmode=log, ymode=log,
    log basis y = 10,
    each nth point={10},
    xmin=1e-4, ymin=1e-4, xmax=1, ymax=1,
    xlabel={FPR}, ylabel={TPR (DTS)},
    grid=major,
    legend pos={south east},
    legend style={/tikz/every node/.append style={anchor=west}},
    every axis plot/.append style={line width=1pt},
    cycle list name=mycolors,
  ]
    
    \addplot [gray, dashed, domain=1e-4:1, samples=100, forget plot] {x};

    \addplot+[
    mark=none]
    table[x=fpr, y=tpr, col sep=comma] {data/EEG/DTS/LSTM-online/total/roc_InceptionTimeS.csv};
    \addlegendentry{InceptionTime}
    
    \addplot+[mark=none]
      table[x=fpr, y=tpr, col sep=comma] {data/EEG/DTS/LSTM-online/total/roc_LSTM.csv};
    \addlegendentry{LSTM}

  \end{axis}
\end{tikzpicture}
\end{subfigure}
\begin{subfigure}[t]{0.22\textwidth}
\centering
\begin{tikzpicture}[scale=0.5]
  \begin{axis}[
    xmode=log, ymode=log,
    xmin=1e-4, ymin=1e-4, xmax=1, ymax=1,
    xlabel={FPR}, ylabel={},
    grid=major,
    each nth point={10},
    legend pos={south east},
    legend style={/tikz/every node/.append style={anchor=west}},
    every axis plot/.append style={line width=1pt},
    cycle list name=mycolors,
  ]
  \addplot [gray, dashed, domain=1e-4:1, samples=100, forget plot] {x};
  
    \addplot+[mark=none]
      table[x=fpr, y=tpr, col sep=comma] {data/EEG/DTS/NHiTS-online/total/roc_InceptionTimeS.csv};
    \addlegendentry{InceptionTime}
    
    \addplot+[mark=none]
      table[x=fpr, y=tpr, col sep=comma] {data/EEG/DTS/NHiTS-online/total/roc_LSTM.csv};
    \addlegendentry{LSTM}
    
  \end{axis}
\end{tikzpicture}
\end{subfigure}
\begin{subfigure}[t]{0.22\textwidth}
\centering
\begin{tikzpicture}[scale=0.5]
  \begin{axis}[
    xmode=log, ymode=log,
    xmin=1e-4, ymin=1e-4, xmax=1, ymax=1,
    xlabel={FPR}, ylabel={},
    grid=major,
    each nth point={10},
    legend pos={south east},
    legend style={/tikz/every node/.append style={anchor=west}},
    every axis plot/.append style={line width=1pt},
    cycle list name=mycolors,
  ]
  \addplot [gray, dashed, domain=1e-4:1, samples=100, forget plot] {x};
  
    \addplot+[mark=none]
      table[x=fpr, y=tpr, col sep=comma] {data/ELD/DTS/LSTM-online/total/roc_InceptionTimeS.csv};
    \addlegendentry{InceptionTime}
    
    \addplot+[mark=none]
      table[x=fpr, y=tpr, col sep=comma] {data/ELD/DTS/LSTM-online/total/roc_LSTM.csv};
    \addlegendentry{LSTM}
    
  \end{axis}
\end{tikzpicture}
\end{subfigure}
\begin{subfigure}[t]{0.22\textwidth}
\centering
\begin{tikzpicture}[scale=0.5]
  \begin{axis}[
    xmode=log, ymode=log,
    xmin=1e-4, ymin=1e-4, xmax=1, ymax=1,
    xlabel={FPR}, ylabel={},
    grid=major,
    each nth point={10},
    legend pos={south east},
    legend style={/tikz/every node/.append style={anchor=west}},
    every axis plot/.append style={line width=1pt},
    cycle list name=mycolors,
  ]
  \addplot [gray, dashed, domain=1e-4:1, samples=100, forget plot] {x};
  
    \addplot+[mark=none]
      table[x=fpr, y=tpr, col sep=comma] {data/ELD/DTS/NHiTS-online/total/roc_InceptionTimeS.csv};
    \addlegendentry{InceptionTime}
    
    \addplot+[mark=none]
      table[x=fpr, y=tpr, col sep=comma] {data/ELD/DTS/NHiTS-online/total/roc_LSTM.csv};
    \addlegendentry{LSTM}

  \end{axis}
\end{tikzpicture}
\end{subfigure}

%% file: attack_ROC_offline_fig.tex
\pgfplotscreateplotcyclelist{mycolors}{
    {blue, mark=none},
    {red, mark=none},
    {green!60!black, mark=none},
    {orange, mark=none},
    {purple, mark=none},
    {brown, mark=none},
    {cyan!60!black, mark=none},
}

\begin{figure*}[ht]
\centering

\input{offline_ROCs_rows/RMIA}

\input{offline_ROCs_rows/LiRA}

 \input{offline_ROCs_rows/Multi-LiRA}

 \input{offline_ROCs_rows/DTS}

\par\vspace{1em}

\caption{ROC curves with logarithmic scales (TPR on the y-axis and FPR on the x-axis) for four memberships inference attacks: RMIA, LiRA, Multi-Signal LiRA, and DTS, arranged in columns. Rows correspond to dataset–architecture pairs: EEG–LSTM, EEG–N-HiTS, ELD–LSTM, and ELD–N-HiTS. 
All results are for the \textbf{offline} setting. Within each subplot, curves correspond to individual attack signals for LiRA and RMIA, all combined signals for Multi-Signal LiRA, and different classifier architectures for DTS. Each curve represents the mean over five independent runs.
}
\label{fig:roc_offline}
\end{figure*}
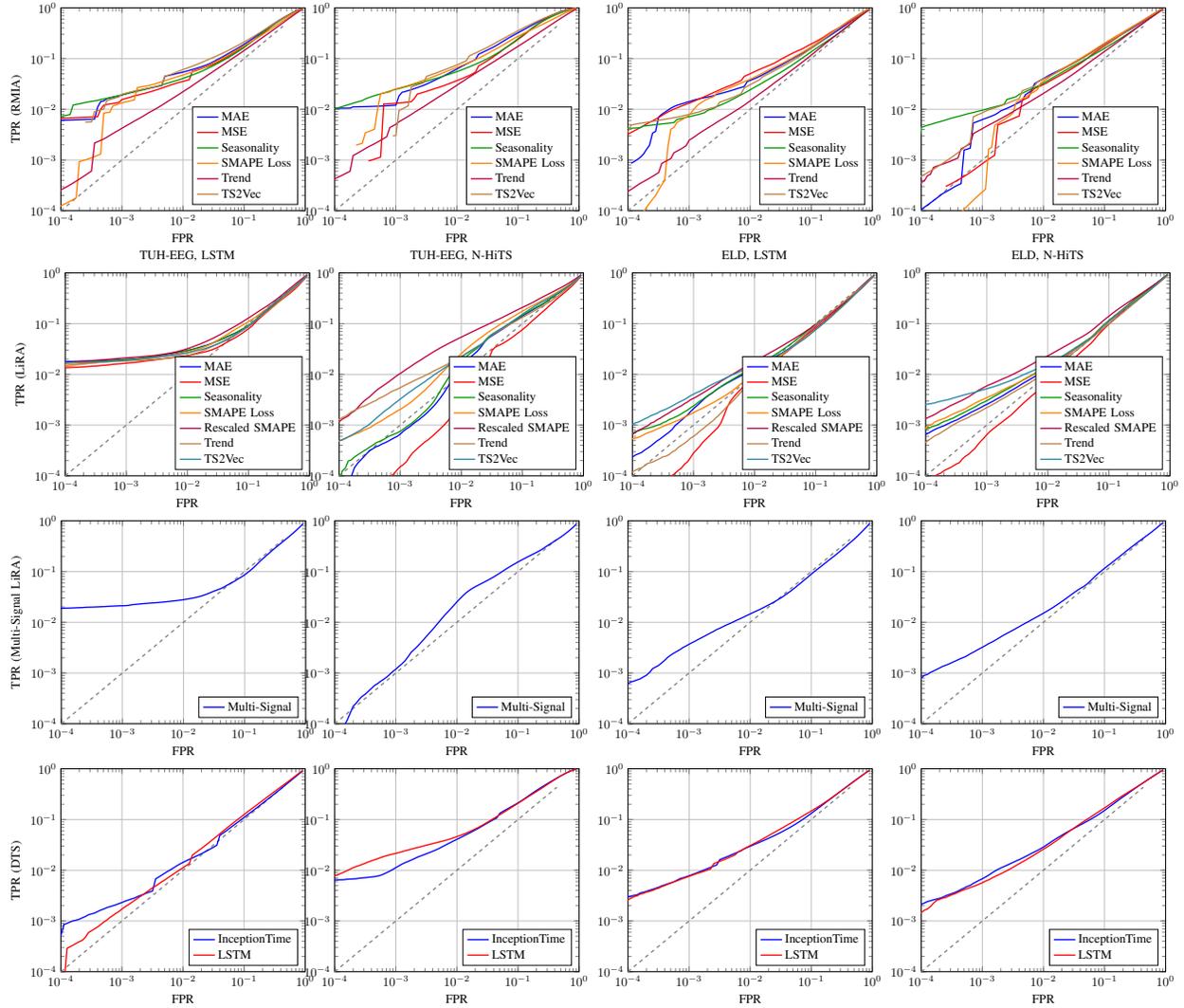

%% file: offline_ROCs_rows/RMIA.tex
\pgfplotsset{
  every axis/.append style={
    label style={font=\normalsize},
    tick label style={font=\normalsize},
    legend style={font=\normalsize},
  }
}

\begin{subfigure}[t]{0.22\textwidth}
\centering
\begin{tikzpicture}[scale=0.5]
  \begin{axis}[
    xmode=log, ymode=log,
    log basis y = 10,
    each nth point={10},
    xmin=1e-4, ymin=1e-4, xmax=1, ymax=1,
    xlabel={FPR}, ylabel={TPR (RMIA)},
    grid=major,
    legend pos={south east},
    legend style={/tikz/every node/.append style={anchor=west}},
    every axis plot/.append style={line width=1pt},
    cycle list name=mycolors,
  ]
    
    \addplot [gray, dashed, domain=1e-4:1, samples=100, forget plot] {x};

    \addplot+[
    mark=none]
    table[x=fpr, y=tpr, col sep=comma] {data/EEG/RMIA/LSTM-offline/total/roc_MAELoss.csv};
    \addlegendentry{MAE}
    
    \addplot+[mark=none]
      table[x=fpr, y=tpr, col sep=comma] {data/EEG/RMIA/LSTM-offline/total/roc_MSELoss.csv};
    \addlegendentry{MSE}
    
    \addplot+[mark=none]
      table[x=fpr, y=tpr, col sep=comma] {data/EEG/RMIA/LSTM-offline/total/roc_SeasonalityLoss.csv};
    \addlegendentry{Seasonality}
    
    \addplot+[mark=none]
      table[x=fpr, y=tpr, col sep=comma] {data/EEG/RMIA/LSTM-offline/total/roc_SMAPELoss.csv};
    \addlegendentry{SMAPE Loss}
    
    \addplot+[mark=none]
      table[x=fpr, y=tpr, col sep=comma] {data/EEG/RMIA/LSTM-offline/total/roc_TrendLoss.csv};
    \addlegendentry{Trend}
    
    \addplot+[mark=none]
      table[x=fpr, y=tpr, col sep=comma] {data/EEG/RMIA/LSTM-offline/total/roc_TS2VecLoss.csv};
    \addlegendentry{TS2Vec}
    
  \end{axis}
\end{tikzpicture}
\end{subfigure}
\begin{subfigure}[t]{0.22\textwidth}
\centering
\begin{tikzpicture}[scale=0.5]
  \begin{axis}[
    xmode=log, ymode=log,
    xmin=1e-4, ymin=1e-4, xmax=1, ymax=1,
    xlabel={FPR}, ylabel={},
    grid=major,
    each nth point={10},
    legend pos={south east},
    legend style={/tikz/every node/.append style={anchor=west}},
    every axis plot/.append style={line width=1pt},
    cycle list name=mycolors,
  ]
  \addplot [gray, dashed, domain=1e-4:1, samples=100, forget plot] {x};
  
    \addplot+[mark=none]
      table[x=fpr, y=tpr, col sep=comma] {data/EEG/RMIA/NHiTS-offline/total/roc_MAELoss.csv};
    \addlegendentry{MAE}
    
    \addplot+[mark=none]
      table[x=fpr, y=tpr, col sep=comma] {data/EEG/RMIA/NHiTS-offline/total/roc_MSELoss.csv};
    \addlegendentry{MSE}
    
    \addplot+[mark=none]
      table[x=fpr, y=tpr, col sep=comma] {data/EEG/RMIA/NHiTS-offline/total/roc_SeasonalityLoss.csv};
    \addlegendentry{Seasonality}
    
    \addplot+[mark=none]
      table[x=fpr, y=tpr, col sep=comma] {data/EEG/RMIA/NHiTS-offline/total/roc_SMAPELoss.csv};
    \addlegendentry{SMAPE Loss}
    
    \addplot+[mark=none]
      table[x=fpr, y=tpr, col sep=comma] {data/EEG/RMIA/NHiTS-offline/total/roc_TrendLoss.csv};
    \addlegendentry{Trend}
    
    \addplot+[mark=none]
      table[x=fpr, y=tpr, col sep=comma] {data/EEG/RMIA/NHiTS-offline/total/roc_TS2VecLoss.csv};
    \addlegendentry{TS2Vec}
  \end{axis}
\end{tikzpicture}
\end{subfigure}
\begin{subfigure}[t]{0.22\textwidth}
\centering
\begin{tikzpicture}[scale=0.5]
  \begin{axis}[
    xmode=log, ymode=log,
    xmin=1e-4, ymin=1e-4, xmax=1, ymax=1,
    xlabel={FPR}, ylabel={},
    grid=major,
    each nth point={10},
    legend pos={south east},
    legend style={/tikz/every node/.append style={anchor=west}},
    every axis plot/.append style={line width=1pt},
    cycle list name=mycolors,
  ]
  \addplot [gray, dashed, domain=1e-4:1, samples=100, forget plot] {x};
  
    \addplot+[mark=none]
      table[x=fpr, y=tpr, col sep=comma] {data/ELD/RMIA/LSTM-offline/total/roc_MAELoss.csv};
    \addlegendentry{MAE}
    
    \addplot+[mark=none]
      table[x=fpr, y=tpr, col sep=comma] {data/ELD/RMIA/LSTM-offline/total/roc_MSELoss.csv};
    \addlegendentry{MSE}
    
    \addplot+[mark=none]
      table[x=fpr, y=tpr, col sep=comma] {data/ELD/RMIA/LSTM-offline/total/roc_SeasonalityLoss.csv};
    \addlegendentry{Seasonality}
    
    \addplot+[mark=none]
      table[x=fpr, y=tpr, col sep=comma] {data/ELD/RMIA/LSTM-offline/total/roc_SMAPELoss.csv};
    \addlegendentry{SMAPE Loss}
    
    \addplot+[mark=none]
      table[x=fpr, y=tpr, col sep=comma] {data/ELD/RMIA/LSTM-offline/total/roc_TrendLoss.csv};
    \addlegendentry{Trend}
    
    \addplot+[mark=none]
      table[x=fpr, y=tpr, col sep=comma] {data/ELD/RMIA/LSTM-offline/total/roc_TS2VecLoss.csv};
    \addlegendentry{TS2Vec}
  \end{axis}
\end{tikzpicture}
\end{subfigure}
\begin{subfigure}[t]{0.22\textwidth}
\centering
\begin{tikzpicture}[scale=0.5]
  \begin{axis}[
    xmode=log, ymode=log,
    xmin=1e-4, ymin=1e-4, xmax=1, ymax=1,
    xlabel={FPR}, ylabel={},
    grid=major,
    each nth point={10},
    legend pos={south east},
    legend style={/tikz/every node/.append style={anchor=west}},
    every axis plot/.append style={line width=1pt},
    cycle list name=mycolors,
  ]
  \addplot [gray, dashed, domain=1e-4:1, samples=100, forget plot] {x};
  
    \addplot+[mark=none]
      table[x=fpr, y=tpr, col sep=comma] {data/ELD/RMIA/NHiTS-offline/total/roc_MAELoss.csv};
    \addlegendentry{MAE}
    
    \addplot+[mark=none]
      table[x=fpr, y=tpr, col sep=comma] {data/ELD/RMIA/NHiTS-offline/total/roc_MSELoss.csv};
    \addlegendentry{MSE}
    
    \addplot+[mark=none]
      table[x=fpr, y=tpr, col sep=comma] {data/ELD/RMIA/NHiTS-offline/total/roc_SeasonalityLoss.csv};
    \addlegendentry{Seasonality}
    
    \addplot+[mark=none]
      table[x=fpr, y=tpr, col sep=comma] {data/ELD/RMIA/NHiTS-offline/total/roc_SMAPELoss.csv};
    \addlegendentry{SMAPE Loss}
    
    \addplot+[mark=none]
      table[x=fpr, y=tpr, col sep=comma] {data/ELD/RMIA/NHiTS-offline/total/roc_TrendLoss.csv};
    \addlegendentry{Trend}
    
    \addplot+[mark=none]
      table[x=fpr, y=tpr, col sep=comma] {data/ELD/RMIA/NHiTS-offline/total/roc_TS2VecLoss.csv};
    \addlegendentry{TS2Vec}
  \end{axis}
\end{tikzpicture}
\end{subfigure}

%% file: offline_ROCs_rows/LiRA.tex
\pgfplotsset{
  every axis/.append style={
    label style={font=\normalsize},
    tick label style={font=\normalsize},
    legend style={font=\normalsize},
  }
}

\begin{subfigure}[t]{0.22\textwidth}
\centering
\begin{tikzpicture}[scale=0.5]
  \begin{axis}[
    xmode=log, ymode=log,
    log basis y = 10,
    each nth point={10},
    xmin=1e-4, ymin=1e-4, xmax=1, ymax=1,
    xlabel={FPR}, ylabel={TPR (LiRA)},
    grid=major,
    legend pos={south east},
    legend style={/tikz/every node/.append style={anchor=west}},
    every axis plot/.append style={line width=1pt},
    cycle list name=mycolors,
     title={TUH-EEG, LSTM},
  ]
    
    \addplot [gray, dashed, domain=1e-4:1, samples=100, forget plot] {x};

    \addplot+[
    mark=none]
    table[x=fpr, y=tpr, col sep=comma] {data/EEG/LiRA/LSTM-offline/total/roc_MAELoss.csv};
    \addlegendentry{MAE}
    
    \addplot+[mark=none]
      table[x=fpr, y=tpr, col sep=comma] {data/EEG/LiRA/LSTM-offline/total/roc_MSELoss.csv};
    \addlegendentry{MSE}
    
    \addplot+[mark=none]
      table[x=fpr, y=tpr, col sep=comma] {data/EEG/LiRA/LSTM-offline/total/roc_SeasonalityLoss.csv};
    \addlegendentry{Seasonality}
    
    \addplot+[mark=none]
      table[x=fpr, y=tpr, col sep=comma] {data/EEG/LiRA/LSTM-offline/total/roc_SMAPELoss.csv};
    \addlegendentry{SMAPE Loss}
    
    \addplot+[mark=none]
      table[x=fpr, y=tpr, col sep=comma] {data/EEG/LiRA/LSTM-offline/total/roc_RescaledSMAPELoss.csv};
    \addlegendentry{Rescaled SMAPE}
    
    \addplot+[mark=none]
      table[x=fpr, y=tpr, col sep=comma] {data/EEG/LiRA/LSTM-offline/total/roc_TrendLoss.csv};
    \addlegendentry{Trend}
    
    \addplot+[mark=none]
      table[x=fpr, y=tpr, col sep=comma] {data/EEG/LiRA/LSTM-offline/total/roc_TS2VecLoss.csv};
    \addlegendentry{TS2Vec}
    
  \end{axis}
\end{tikzpicture}
\end{subfigure}
\begin{subfigure}[t]{0.22\textwidth}
\centering
\begin{tikzpicture}[scale=0.5]
  \begin{axis}[
    xmode=log, ymode=log,
    xmin=1e-4, ymin=1e-4, xmax=1, ymax=1,
    xlabel={FPR}, ylabel={},
    grid=major,
    each nth point={10},
    legend pos={south east},
    legend style={/tikz/every node/.append style={anchor=west}},
    every axis plot/.append style={line width=1pt},
    cycle list name=mycolors,
     title={TUH-EEG, N-HiTS},
  ]
  \addplot [gray, dashed, domain=1e-4:1, samples=100, forget plot] {x};
  
    \addplot+[mark=none]
      table[x=fpr, y=tpr, col sep=comma] {data/EEG/LiRA/NHiTS-offline/total/roc_MAELoss.csv};
    \addlegendentry{MAE}
    
    \addplot+[mark=none]
      table[x=fpr, y=tpr, col sep=comma] {data/EEG/LiRA/NHiTS-offline/total/roc_MSELoss.csv};
    \addlegendentry{MSE}
    
    \addplot+[mark=none]
      table[x=fpr, y=tpr, col sep=comma] {data/EEG/LiRA/NHiTS-offline/total/roc_SeasonalityLoss.csv};
    \addlegendentry{Seasonality}
    
    \addplot+[mark=none]
      table[x=fpr, y=tpr, col sep=comma] {data/EEG/LiRA/NHiTS-offline/total/roc_SMAPELoss.csv};
    \addlegendentry{SMAPE Loss}
    
    \addplot+[mark=none]
      table[x=fpr, y=tpr, col sep=comma] {data/EEG/LiRA/NHiTS-offline/total/roc_RescaledSMAPELoss.csv};
    \addlegendentry{Rescaled SMAPE}
    
    \addplot+[mark=none]
      table[x=fpr, y=tpr, col sep=comma] {data/EEG/LiRA/NHiTS-offline/total/roc_TrendLoss.csv};
    \addlegendentry{Trend}
    
    \addplot+[mark=none]
      table[x=fpr, y=tpr, col sep=comma] {data/EEG/LiRA/NHiTS-offline/total/roc_TS2VecLoss.csv};
    \addlegendentry{TS2Vec}
  \end{axis}
\end{tikzpicture}
\end{subfigure}
\begin{subfigure}[t]{0.22\textwidth}
\centering
\begin{tikzpicture}[scale=0.5]
  \begin{axis}[
    xmode=log, ymode=log,
    xmin=1e-4, ymin=1e-4, xmax=1, ymax=1,
    xlabel={FPR}, ylabel={},
    grid=major,
    each nth point={10},
    legend pos={south east},
    legend style={/tikz/every node/.append style={anchor=west}},
    every axis plot/.append style={line width=1pt},
    cycle list name=mycolors,
     title={ELD, LSTM},
  ]
  \addplot [gray, dashed, domain=1e-4:1, samples=100, forget plot] {x};
  
    \addplot+[mark=none]
      table[x=fpr, y=tpr, col sep=comma] {data/ELD/LiRA/LSTM-offline/total/roc_MAELoss.csv};
    \addlegendentry{MAE}
    
    \addplot+[mark=none]
      table[x=fpr, y=tpr, col sep=comma] {data/ELD/LiRA/LSTM-offline/total/roc_MSELoss.csv};
    \addlegendentry{MSE}
    
    \addplot+[mark=none]
      table[x=fpr, y=tpr, col sep=comma] {data/ELD/LiRA/LSTM-offline/total/roc_SeasonalityLoss.csv};
    \addlegendentry{Seasonality}
    
    \addplot+[mark=none]
      table[x=fpr, y=tpr, col sep=comma] {data/ELD/LiRA/LSTM-offline/total/roc_SMAPELoss.csv};
    \addlegendentry{SMAPE Loss}
    
    \addplot+[mark=none]
      table[x=fpr, y=tpr, col sep=comma] {data/ELD/LiRA/LSTM-offline/total/roc_RescaledSMAPELoss.csv};
    \addlegendentry{Rescaled SMAPE}
    
    \addplot+[mark=none]
      table[x=fpr, y=tpr, col sep=comma] {data/ELD/LiRA/LSTM-offline/total/roc_TrendLoss.csv};
    \addlegendentry{Trend}
    
    \addplot+[mark=none]
      table[x=fpr, y=tpr, col sep=comma] {data/ELD/LiRA/LSTM-offline/total/roc_TS2VecLoss.csv};
    \addlegendentry{TS2Vec}
  \end{axis}
\end{tikzpicture}
\end{subfigure}
\begin{subfigure}[t]{0.22\textwidth}
\centering
\begin{tikzpicture}[scale=0.5]
  \begin{axis}[
    xmode=log, ymode=log,
    xmin=1e-4, ymin=1e-4, xmax=1, ymax=1,
    xlabel={FPR}, ylabel={},
    grid=major,
    each nth point={10},
    legend pos={south east},
    legend style={/tikz/every node/.append style={anchor=west}},
    every axis plot/.append style={line width=1pt},
    cycle list name=mycolors,
     title={ELD, N-HiTS},
  ]
  \addplot [gray, dashed, domain=1e-4:1, samples=100, forget plot] {x};
  
    \addplot+[mark=none]
      table[x=fpr, y=tpr, col sep=comma] {data/ELD/LiRA/NHiTS-offline/total/roc_MAELoss.csv};
    \addlegendentry{MAE}
    
    \addplot+[mark=none]
      table[x=fpr, y=tpr, col sep=comma] {data/ELD/LiRA/NHiTS-offline/total/roc_MSELoss.csv};
    \addlegendentry{MSE}
    
    \addplot+[mark=none]
      table[x=fpr, y=tpr, col sep=comma] {data/ELD/LiRA/NHiTS-offline/total/roc_SeasonalityLoss.csv};
    \addlegendentry{Seasonality}
    
    \addplot+[mark=none]
      table[x=fpr, y=tpr, col sep=comma] {data/ELD/LiRA/NHiTS-offline/total/roc_SMAPELoss.csv};
    \addlegendentry{SMAPE Loss}
    
    \addplot+[mark=none]
      table[x=fpr, y=tpr, col sep=comma] {data/ELD/LiRA/NHiTS-offline/total/roc_RescaledSMAPELoss.csv};
    \addlegendentry{Rescaled SMAPE}
    
    \addplot+[mark=none]
      table[x=fpr, y=tpr, col sep=comma] {data/ELD/LiRA/NHiTS-offline/total/roc_TrendLoss.csv};
    \addlegendentry{Trend}
    
    \addplot+[mark=none]
      table[x=fpr, y=tpr, col sep=comma] {data/ELD/LiRA/NHiTS-offline/total/roc_TS2VecLoss.csv};
    \addlegendentry{TS2Vec}
  \end{axis}
\end{tikzpicture}
\end{subfigure}

%% file: offline_ROCs_rows/Multi-LiRA.tex
\pgfplotsset{
  every axis/.append style={
    label style={font=\normalsize},
    tick label style={font=\normalsize},
    legend style={font=\normalsize},
  }
}

\begin{subfigure}[t]{0.22\textwidth}
\centering
\begin{tikzpicture}[scale=0.5]
  \begin{axis}[
    xmode=log, ymode=log,
    log basis y = 10,
    each nth point={10},
    xmin=1e-4, ymin=1e-4, xmax=1, ymax=1,
    xlabel={FPR}, ylabel={TPR (Multi-Signal LiRA)},
    grid=major,
    legend pos={south east},
    legend style={/tikz/every node/.append style={anchor=west}},
    every axis plot/.append style={line width=1pt},
    cycle list name=mycolors,
  ]
    
    \addplot [gray, dashed, domain=1e-4:1, samples=100, forget plot] {x};

    \addplot+[
    mark=none]
    table[x=fpr, y=tpr, col sep=comma] {data/EEG/Multi-LiRA/LSTM-offline/total/roc.csv};
    \addlegendentry{Multi-Signal}
    
  \end{axis}
\end{tikzpicture}
\end{subfigure}
\begin{subfigure}[t]{0.22\textwidth}
\centering
\begin{tikzpicture}[scale=0.5]
  \begin{axis}[
    xmode=log, ymode=log,
    xmin=1e-4, ymin=1e-4, xmax=1, ymax=1,
    xlabel={FPR}, ylabel={},
    grid=major,
    each nth point={10},
    legend pos={south east},
    legend style={/tikz/every node/.append style={anchor=west}},
    every axis plot/.append style={line width=1pt},
    cycle list name=mycolors,
  ]
  \addplot [gray, dashed, domain=1e-4:1, samples=100, forget plot] {x};
  
    \addplot+[mark=none]
      table[x=fpr, y=tpr, col sep=comma] {data/EEG/Multi-LiRA/NHiTS-offline/total/roc.csv};
    \addlegendentry{Multi-Signal}
    
  \end{axis}
\end{tikzpicture}
\end{subfigure}
\begin{subfigure}[t]{0.22\textwidth}
\centering
\begin{tikzpicture}[scale=0.5]
  \begin{axis}[
    xmode=log, ymode=log,
    xmin=1e-4, ymin=1e-4, xmax=1, ymax=1,
    xlabel={FPR}, ylabel={},
    grid=major,
    each nth point={10},
    legend pos={south east},
    legend style={/tikz/every node/.append style={anchor=west}},
    every axis plot/.append style={line width=1pt},
    cycle list name=mycolors,
  ]
  \addplot [gray, dashed, domain=1e-4:1, samples=100, forget plot] {x};
  
    \addplot+[mark=none]
      table[x=fpr, y=tpr, col sep=comma] {data/ELD/Multi-LiRA/LSTM-offline/total/roc.csv};
    \addlegendentry{Multi-Signal}
  \end{axis}
\end{tikzpicture}
\end{subfigure}
\begin{subfigure}[t]{0.22\textwidth}
\centering
\begin{tikzpicture}[scale=0.5]
  \begin{axis}[
    xmode=log, ymode=log,
    xmin=1e-4, ymin=1e-4, xmax=1, ymax=1,
    xlabel={FPR}, ylabel={},
    grid=major,
    each nth point={10},
    legend pos={south east},
    legend style={/tikz/every node/.append style={anchor=west}},
    every axis plot/.append style={line width=1pt},
    cycle list name=mycolors,
  ]
  \addplot [gray, dashed, domain=1e-4:1, samples=100, forget plot] {x};
  
    \addplot+[mark=none]
      table[x=fpr, y=tpr, col sep=comma] {data/ELD/Multi-LiRA/NHiTS-offline/total/roc.csv};
    \addlegendentry{Multi-Signal}
    
  \end{axis}
\end{tikzpicture}
\end{subfigure}

%% file: offline_ROCs_rows/DTS.tex
\pgfplotsset{
  every axis/.append style={
    label style={font=\normalsize},
    tick label style={font=\normalsize},
    legend style={font=\normalsize},
  }
}

\begin{subfigure}[t]{0.22\textwidth}
\centering
\begin{tikzpicture}[scale=0.5]
  \begin{axis}[
    xmode=log, ymode=log,
    log basis y = 10,
    each nth point={10},
    xmin=1e-4, ymin=1e-4, xmax=1, ymax=1,
    xlabel={FPR}, ylabel={TPR (DTS)},
    grid=major,
    legend pos={south east},
    legend style={/tikz/every node/.append style={anchor=west}},
    every axis plot/.append style={line width=1pt},
    cycle list name=mycolors,
  ]
    
    \addplot [gray, dashed, domain=1e-4:1, samples=100, forget plot] {x};

    \addplot+[
    mark=none]
    table[x=fpr, y=tpr, col sep=comma] {data/EEG/DTS/LSTM-offline/total/roc_InceptionTimeS.csv};
    \addlegendentry{InceptionTime}
    
    \addplot+[mark=none]
      table[x=fpr, y=tpr, col sep=comma] {data/EEG/DTS/LSTM-offline/total/roc_LSTM.csv};
    \addlegendentry{LSTM}

  \end{axis}
\end{tikzpicture}
\end{subfigure}
\begin{subfigure}[t]{0.22\textwidth}
\centering
\begin{tikzpicture}[scale=0.5]
  \begin{axis}[
    xmode=log, ymode=log,
    xmin=1e-4, ymin=1e-4, xmax=1, ymax=1,
    xlabel={FPR}, ylabel={},
    grid=major,
    each nth point={10},
    legend pos={south east},
    legend style={/tikz/every node/.append style={anchor=west}},
    every axis plot/.append style={line width=1pt},
    cycle list name=mycolors,
  ]
  \addplot [gray, dashed, domain=1e-4:1, samples=100, forget plot] {x};
  
    \addplot+[mark=none]
      table[x=fpr, y=tpr, col sep=comma] {data/EEG/DTS/NHiTS-offline/total/roc_InceptionTimeS.csv};
    \addlegendentry{InceptionTime}
    
    \addplot+[mark=none]
      table[x=fpr, y=tpr, col sep=comma] {data/EEG/DTS/NHiTS-offline/total/roc_LSTM.csv};
    \addlegendentry{LSTM}
    
  \end{axis}
\end{tikzpicture}
\end{subfigure}
\begin{subfigure}[t]{0.22\textwidth}
\centering
\begin{tikzpicture}[scale=0.5]
  \begin{axis}[
    xmode=log, ymode=log,
    xmin=1e-4, ymin=1e-4, xmax=1, ymax=1,
    xlabel={FPR}, ylabel={},
    grid=major,
    each nth point={10},
    legend pos={south east},
    legend style={/tikz/every node/.append style={anchor=west}},
    every axis plot/.append style={line width=1pt},
    cycle list name=mycolors,
  ]
  \addplot [gray, dashed, domain=1e-4:1, samples=100, forget plot] {x};
  
    \addplot+[mark=none]
      table[x=fpr, y=tpr, col sep=comma] {data/ELD/DTS/LSTM-offline/total/roc_InceptionTimeS.csv};
    \addlegendentry{InceptionTime}
    
    \addplot+[mark=none]
      table[x=fpr, y=tpr, col sep=comma] {data/ELD/DTS/LSTM-offline/total/roc_LSTM.csv};
    \addlegendentry{LSTM}
    
  \end{axis}
\end{tikzpicture}
\end{subfigure}
\begin{subfigure}[t]{0.22\textwidth}
\centering
\begin{tikzpicture}[scale=0.5]
  \begin{axis}[
    xmode=log, ymode=log,
    xmin=1e-4, ymin=1e-4, xmax=1, ymax=1,
    xlabel={FPR}, ylabel={},
    grid=major,
    each nth point={10},
    legend pos={south east},
    legend style={/tikz/every node/.append style={anchor=west}},
    every axis plot/.append style={line width=1pt},
    cycle list name=mycolors,
  ]
  \addplot [gray, dashed, domain=1e-4:1, samples=100, forget plot] {x};
  
    \addplot+[mark=none]
      table[x=fpr, y=tpr, col sep=comma] {data/ELD/DTS/NHiTS-offline/total/roc_InceptionTimeS.csv};
    \addlegendentry{InceptionTime}
    
    \addplot+[mark=none]
      table[x=fpr, y=tpr, col sep=comma] {data/ELD/DTS/NHiTS-offline/total/roc_LSTM.csv};
    \addlegendentry{LSTM}

  \end{axis}
\end{tikzpicture}
\end{subfigure}